\documentclass{article}

\usepackage{arxiv}
\usepackage{natbib}
\usepackage{times}
\usepackage{soul}
\usepackage{url}
\usepackage[hidelinks]{hyperref}
\usepackage[utf8]{inputenc}
\usepackage[small]{caption}
\usepackage{graphicx}
\usepackage{amsmath}
\usepackage{amsthm}
\usepackage{booktabs}
\usepackage{algorithm}
\usepackage{algorithmic}
\usepackage[utf8]{inputenc} 
\usepackage[T1]{fontenc}    
\usepackage{amsfonts}       
\usepackage{nicefrac}       
\usepackage{microtype}      
\usepackage{subfigure}
\usepackage{amssymb}

\usepackage{subfiles}
\usepackage{float}
\usepackage{enumerate}
\usepackage{multirow}
\usepackage{booktabs}
\graphicspath{ images/ }
\usepackage{todonotes}
\usepackage{stmaryrd}
\usepackage{comment}
\usepackage{cleveref}
\usepackage{wrapfig}
\usepackage{xspace}
\usepackage{bbm}

\newcommand{\ElOE}{\textsf{ElOE}\xspace}
\newcommand{\triple}[1]{\ensuremath\langle #1\rangle}
\newtheorem{definition}{Definition}
\DeclareMathOperator{\Tr}{Tr}
\title{Elliptical Ordinal Embedding}

\date{} 					

\author{ {A\"issatou Diallo} \\
	Research Training Group AIPHES\\
	Department of Computer Science\\
	Technische Universit\"at Darmstadt \\
	Darmstadt, Germany \\
	\texttt{diallo@aiphes.tu-darmstadt.de}  \\
	\And
	{Johannes F\"urnkranz} \\
	Computational Data Analytics\\
	Institute for Application Oriented Knowledge Processing (FAW)\\
	Johannes Kepler University Linz\\
	Linz, Austria \\
	\texttt{juffi@faw.jku.at} \\
}

\renewcommand{\headeright}{}
\renewcommand{\undertitle}{}

\hypersetup{
pdftitle={Elliptical Ordinal Embedding},
pdfsubject={q-bio.NC, q-bio.QM},
pdfauthor={David S.~Hippocampus, Elias D.~Striatum},
pdfkeywords={First keyword, Second keyword, More},
}

\begin{document}
\maketitle

\begin{abstract}
Ordinal embedding aims at finding a low di\-men\-sional representation of objects from a set of constraints of the form ''item $j$ is closer to item $i$ than item $k$''. 
Typically, each object is mapped  onto a point vector in a low dimensional metric space. We argue that mapping to a density instead of a point vector provides some interesting advantages, including an inherent reflection of the uncertainty about the representation itself and its relative location in the space. Indeed, in this paper, we propose to embed each object as a Gaussian distribution. We investigate the ability of these embeddings to capture the underlying structure of the data while satisfying the constraints, and explore properties of the representation. Experiments on synthetic and real-world datasets showcase the advantages of our approach. In addition, we illustrate the merit of modelling uncertainty, which enriches the visual perception of the mapped objects in the space.
\end{abstract}


\section{Introduction}
\label{sec:introduction}
A crucial problem in machine learning is the assessment of similarities between data instances. In fact, multiple tasks depend on such an ability. For example, in clustering, similar items should be grouped together, or in classification, where similar items should be assigned similar labels. In general, 
one expects to be given a collection of data instances and a similarity function that allows determining how similar objects are to each other. 
Yet, it is not always straight-forward to define such a similarity function for a given data representation.
Thus, recent works in machine learning focus on a scenario 
in which the learner is only given relative comparisons between data instances \citep{agarwal2007generalized,mcfee2012more}.
Instead of directly querying the degree of similarity between items on an absolute scale, it has been shown that eliciting \emph{ordinal feedback} from subjects in the form of ''item $i$ is more similar to item $j$ than to item $k$'' is a more reliable form of supervision, 
especially when the feedback is subjective \citep{stewart2005absolute}. 
The problem of interest is to learn representations in a low-dimensional metric space such that the relative distances of the representation satisfy a set of ordinal triplet constraints of the above type. This problem is known as \textit{ordinal embedding}.
It dates back to the classic non-metric multidimensional scaling approach, but interest in the problem has renewed in recent years.  The main expected result of this task is a faithful geometric representation that allows to easily visualize similarities 
between data instances. 

We argue that the classical representation of items as points does not allow to capture important aspects of the data, such as the inherent noise of the ordinal feedback and the resulting uncertainty of the representation. 
Consider the case of an object for which the different triplets are conflicting with each other or even revealing contradicting underlying patterns. Such a discrepancy should be reflected and possibly visually expressed by the learnt embedding.
As a remedy, we propose 
to embed 
items 
as probability distributions in $\mathbb{R}^d$, with a location and a scale parameter, which captures
the uncertainty of the location. In particular, we focus on Gaussian distributions which enjoy some desirable properties (cf.\ 
\Cref{sec:poe}).

\paragraph{Contributions.} In this paper, we propose a novel ordinal
embedding approach that represents items as Gaussian measures based on ordinal constraints: each object becomes a full distribution rather than a single point. Thereby, we capture the uncertainty about its representation and location in the latent space. Our method draws inspiration from recent developments in optimal transport techniques for representation learning. We adapt these techniques for the ordinal feedback setting using the Wasserstein metric, a distance metric between two probability distributions.
The result is a fast embedding approach which we call \emph{elliptical ordinal embedding} (\ElOE). 
 
We show that \ElOE recovers a latent embedding with a sufficient number of triplets. 
We empirically validate our results on both synthetic and real triplet datasets, and provide examples where elliptical ordinal embedding techniques can be used for a better visualization of complex datasets. 

The paper is organized as follows: \Cref{sec:related_work} reviews related work on ordinal as well as probabilistic embeddings for representation learning. In \Cref{sec:problem}, we formally introduce the problem of ordinal embedding. In \Cref{sec:poe} we present our approach for elliptical ordinal embedding. Finally, \Cref{sec:experiments} presents our empirical studies and analyzes the results.

\section{Related Work}
\label{sec:related_work}


\paragraph{Ordinal Embeddings.} 
In recent years, ordinal data have received a growing interest in machine learning. The ordinal embedding problem has been studied from different points of view: 
Multiple methods have been designed to deal with triplet similarity producing low-dimensional Euclidean vectors as output. In particular, \emph{generalized non-metric multidimensional scaling} (GNMDS) \cite{agarwal2007generalized} relies on a max-margin approach to minimize a hinge loss. \emph{Stochastic triplet embedding} (STE) \cite{van2012stochastic}, on the other hand, assumes a Gaussian noise model and minimizes a logistic noise. The crowd kernel model \citep{ICML-2011-TamuzLBSK} makes the assumption that triplets have been generated by an explicit noise model.  It is worth mentioning that these models adopt a classification scheme to solve the problem by predicting the label of the relative comparisons. \cite{terada2014local} solve the ordinal embedding problem via  reduction to the embedding of nearest-neighbor graphs. Moreover, these methods rely on expensive gradient projections and do not easily scale to large datasets.
The theoretical and statistical guarantees of the ordinal embedding problem have been investigated in \cite{KazemiCDK18,jamieson2011low,jain2016finite}.
%
%
The main purpose of the above methods is to facilitate data visualization of similarity inferred from human assessments. However, other tasks employing similarity triplets have been studied, such as  medoid estimation \citep{heikinheimo2013crowd},  density estimation \citep{ukkonen2015crowdsourced}, or clustering \citep{ukkonen2017crowdsourced}.
Closely related to our approach are \cite{haghiri2019large,AndertonA19}, which employs deep learning to scale the ordinal problem to large datasets. 
\paragraph{Probabilistic Embeddings.} 
The work of \citet{vilnis2014word} established a new trend  in the representation learning field
by proposing
to embed data points, in this case words, as probability distributions in $\mathbb{R}^d$.  Representing objects in the latent space as probability distributions allows more flexibility in the representation and even express multi-modality. 
In fact, point embeddings can be considered as a special case of probabilistic embeddings, namely a Dirac distribution, where the uncertainty is collapsed in a single point.
In the above-mentioned work \citet{vilnis2014word}, the metric used is KL divergence. However, this metric has a drawback: when variances of the probabilistic embedding collapse, the measure does not coincide with the Euclidean metric between point embeddings. Loosely speaking, the KL divergence and the $\ell_2$ distance between two probability measures diverge to infinity when the variances become too small. In addition, the KL divergence does not behave well when the two compared distributions have little or no overlap. It has been shown by \citet{muzellec2018generalizing} that the Wasserstein metric is a better metric to compare probabilistic embeddings.
\citet{ChumbalovMG20} propose to use a distributional embedding based on triplet comparisons with and without features for the task of information retrieval based on the maximization the evidence lower bound.


\section{Problem Statement}
\label{sec:problem}
\newcommand\defeq{\mathrel{\overset{\makebox[0pt]{\mbox{\normalfont\tiny\sffamily def}}}{=}}}


In this section, we formally state the ordinal embedding problem and establish the notation, for which we follow \citet{muzellec2018generalizing}.
%
$\|\cdot\|$ denotes the $\ell_2$ norm.
$\mathcal{S}^{d}_+$ is the set of all positive definite matrices. 
In the scope of this work, we only focus on Gaussian distributions which belong to the family of parametrized probability distributions $z_{h,\mathbf{a}, \mathbf{A}}$ having a location vector $\mathbf{a} \in \mathbb{R}^d$ which represents the shift of the distribution, a scale parameter $\mathbf{A} \in \mathcal{S}^{d}_+ $, which represents the statistical dispersion of the distribution, and a characteristic generator function $h$. Specifically, for Gaussian distributions, the scale parameter coincides with the covariance matrix $var(z_{h,\mathbf{a}, \mathbf{A}}) = \mathbf{A}$. From now on, we denote Gaussian 
distributions (or embeddings) as $z_{(h,\mathbf{a}, \mathbf{A})} = \mathcal{N}(\mathbf{a}, \mathbf{A})$.

Consider $n$ items in an abstract space $\mathcal{X}$, which
we represent by their indices $[n]={1,...,n}$. It is worth mentioning that no explicit representation of the items is available so it is not possible to analytically express the dissimilarity between the items. We assume a latent underlying dissimilarity (or similarity) function $\delta : \mathcal{X} \times \mathcal{X} \xrightarrow{}\mathbb{R}_{\geq 0} $. Let $\mathcal{T} := \{\triple{i,j,k}: 1 \leq i \neq j \neq k \leq n\}$ be a set of unique triplets of elements in $\mathcal{X}$. We further have access to an oracle $\mathcal{O}$ which 
indicates whether 
the inequality 
$\delta(i,j) < \delta(i,k)$  holds or not:

\begin{equation}
\mathcal{O}(\triple{i,j,k}) =  \left\{
    \begin{array}{ll}
        +1 & \mbox{if } \delta(i,j) < \delta(i,k)  \\
        -1 & \mbox{if } \delta(i,j) > \delta(i,k)
    \end{array}
\right.
\label{eq:oracle}
\end{equation}
Note that at this stage, we do not require the latent function $\delta$ to be a metric. 
%
Together, $\mathcal{T}$ and $\mathcal{O}$  represent the observed \emph{ordinal constraints} on distances. 

We can now formally define the 
problem as follows:

\begin{definition}[Ordinal Embedding]
\label{def:classic_ordinal_embedding}
Consider $n$ vector points $\mathbf{X} = (\mathbf{x_1}, \mathbf{x_2}, ..., \mathbf{x_n})$ in a $d$-dimensional Euclidean space $\mathcal{X}$. Given a set of triplets $\mathcal{T}\subset \mathcal{X}^3$ and an oracle $\mathcal{O}: \mathcal{X}^3 \rightarrow \{-1,1\}$,  
the \emph{ordinal embedding problem} consists of recovering $\mathbf{X}$ given $\mathcal{O}$ and $\mathcal{T}$.
\end{definition}

As introduced in \Cref{sec:introduction}, a common application for such an ordinal setting is crowd-sourcing, where many untrained workers complete given tasks. The unequivocal advantage is that this method is a solution for the issue of comparing subjective scales across different crowd-workers.

\section{Elliptical Ordinal Embedding}
\label{sec:poe}
We propose to learn probabilistic embeddings in lieu of the conventional Euclidean embeddings, taking advantage of the fact that vectors can be considered as an extreme case of probability measures, namely a Dirac \citep{muzellec2018generalizing}.  For this purpose, we focus on the family of elliptical distributions, more precisely Gaussian distributions, which enjoy many advantages.
Our goal is to extend the ordinal embedding problem, which we defined in \Cref{def:classic_ordinal_embedding} for Euclidean embeddings, to Gaussian embeddings. 
Hence the considered problem becomes:
\begin{definition}[Probabilistic Ordinal Embedding]
\label{def:probabilistic_ordinal_embedding}
 Suppose $\mathcal{T}\subset \mathcal{X}^3$ is a set of triplets over $\mathcal{X}$ and $\mathcal{O}: \mathcal{X}^3 \rightarrow \{-1,1\}$ is an oracle as defined in \eqref{eq:oracle}. Let $\mathbf{Z}=\{ \mathbf{z}_{1}, \ldots , \mathbf{z}_{n}  \}$ the desired probabilistic embedding, where each of the original points $\mathbf{x}_i$ is mapped to probability distribution  parametrized by $\mathbf{z}_i$. \emph{Probabilistic ordinal embedding} is the problem of obtaining $\mathbf{Z}$ from ordinal constraints $\mathcal{T}$ and $\mathcal{O}$ and a distance measure $d$ such that $\emph{sgn}(d(\mathbf{z}_i,\mathbf{z}_j) - 
 d(\mathbf{z}_i,\mathbf{z}_k))=O(\triple{i,j,k})$, for $\triple{i,j,k} \in \mathcal{T}$.
\end{definition}

This definition requires a distance measure $d$ between distributions.
For this purpose, we selected
the \emph{Wasserstein distance} \citep{olkin1982distance}, also known as \emph{Earth Mover's distance}, which has been previously used as a loss function for supervised learning \citep{frogner2015learning} and in several applications. 

\paragraph{The 2-Wasserstein distance.} In Optimal Transport (OT) theory, the Wasserstein or Kantorovich–Rubinstein metric is a distance function defined between probability distributions (measures) on a given metric space $M$. The squared Wasserstein metric for two arbitrary probability measures $\mu, \nu \in \mathcal{P}(\mathbb{R}^d)$ is defined as: 

\begin{equation*}
    W_2^2(\mu, \nu) \defeq{} \inf_{X\sim \mu, Y\sim \nu } \mathbb{E}_{\|X-Y\|^2}
\end{equation*}
In the general case, it is difficult to find analytical solutions for the Wasserstein distance. However, a closed form solution exists in the case of Gaussian distributions. Let $\alpha \defeq{}\mathcal{N}(\mathbf{a}, \mathbf{A})$ and $\beta \defeq{}\mathcal{N}(\mathbf{b}, \mathbf{B})$, where $\mathbf{a},\mathbf{b} \in \mathbb{R}^d$ and $\mathbf{A}, \mathbf{B} \in \mathcal{S}^d_+$ are positive semi-definite. Hence:
\begin{equation}
\label{eq:wasserstein}
    W^2_2(\alpha, \beta) = \|\mathbf{a}-\mathbf{b}\|^2 + \mathfrak{B}^2(\mathbf{A}, \mathbf{B}) 
\end{equation}
where $\mathfrak{B}^2$ is the \textit{squared Bures metric} \citep{dittmann1999explicit}, defined as:
\begin{equation}
    \mathfrak{B}^2(\mathbf{A}, \mathbf{B}) \defeq{} \text{Tr}(\mathbf{A} + \mathbf{B} -2(\mathbf{A}^{\frac{1}{2}}\mathbf{B}\mathbf{A}^{\frac{1}{2}})^{\frac{1}{2}})
\end{equation}
When $\mathbf{A}=\text{diag }\mathbf{d_A}$ and $\mathbf{B}=\text{diag }\mathbf{d_B}$ are diagonal, $W^2_2$ simplifies to the sum of two terms:
\begin{equation}
     W^2_2(\alpha, \beta) = \|\mathbf{a}-\mathbf{b}\|^2 + \mathfrak{h}^2(\mathbf{d_A}, \mathbf{d_B})
\end{equation}
where $\mathfrak{h}^2(\mathbf{d_A},\mathbf{d_B}) \defeq{} \|\sqrt{\mathbf{d_A}} - \sqrt{\mathbf{d_B}}\|^2$ is the \textit{squared Hellinger distance} \citep{beran1977minimum} between the diagonal $\mathbf{d_A}$ and $\mathbf{d_B}$.

\paragraph{Learning problem.} As mentioned earlier, the goal is to learn a function that maps each item to a $d$-dimensional Gaussian embeddings in ${\mathbb{R}^d}$ such that the 2-Wasserstein distances between the embeddings satisfy as many triplets as possible. Each Gaussian embedding is denoted as $\mathbf{z}_{\mathbf{\mu}, \mathbf{\Sigma}}$, which for the sake of compactness, we abbreviate in $\mathbf{z}$. 
Let $E_{ij}$ be the energy function between two items $(i,j)$ \citep{lecun2006tutorial} which characterizes our energy-based learning approach. In particular, we set $E_{ij}= W^2_2(\mathbf{z}_i,\mathbf{z}_j)$. Finally, the corresponding optimization problem is the following:
\begin{equation}
    \max_{\mathbf{z}_1, ...\mathbf{z}_n \in \mathbb{R}^d}\sum_{t=(i,j,k)\in \mathcal{T}} \mathcal{O}(t)\cdot\text{sgn}(E_{ij} - E_{ik})
\end{equation}
which is discrete, non-convex and NP-hard. 
For these reasons, a relaxation of this optimization problem is needed. We make the choice of using the hinge loss $\mathcal{L}((t=\triple{i,j,k}, \mathcal{O}(t))$, a well established loss function in contrastive metric learning, as a convex surrogate:
\begin{equation}
    \mathcal{L}=\sum_{t=\triple{i,j,k}\in \mathcal{T}} \max(1 - \mathcal{O}(t)
    \cdot 
    (E_{ij} - E_{ik}), 0)
    \label{eq:loss_func}
\end{equation}
The empirical performance of embedding methods is evaluated by the \textit{empirical error}. 
\begin{equation}
    Err = \frac{1}{|T'|}\sum_{\triple{i,j,k} \in \mathcal{T}}\mathbbm{1}[(y\cdot \text{sgn}(E_{ij} - E_{ik})) = 1]  
\end{equation}

\paragraph{Deep neural encoder.} We now detail how the parameters of the elliptical embedding are learnt. 
While our work relates to numerous architectures proposed 
in metric learning such as \citep{hoffer2015deep}, our deep neural encoder have a fundamental difference due to the nature of the problem. The most distinctive point is that we do not have access to the features of the items we aim to embed. In fact our model learns a representation of the items based on a random input to the encoder. In particular, we chose as inputs random vectors on input dimension $h=50$ sampled from $\mathcal{N}(0,I_h)$. A first deep encoder $f_\theta(\cdot)$ processes these random inputs and outputs a representation that is then fed to $\mu_\theta(\cdot)$ and $\Sigma_\theta(\cdot)$, two functions that are non-linear deep-forward networks parametrized by $\theta$ as well. 

\paragraph{Learning.} Besides relying on the optimization of the energy-based max-margin loss \eqref{eq:loss_func}, we apply some regularization to the learning process. We observed that no regularization is needed for learning the location vectors. However, the covariance matrix needs to be bounded, since the main goal of our approach is to obtain perceptual embeddings. Hence, we constrained the covariance matrix to lie within the hypercube $[0,C]^d$, $C$ being a chosen constant.
We chose to focus on diagonal covariance because we argue the rotation angle is not easily interpretable to appreciate the similarity between items and that the principal axes are sufficient to appreciate the uncertainty of the representation. Thus, the regularization is achieved by bounding each element of the covariance matrix, $\Sigma_{ii} = \max(\Sigma_{ii}, C)$ (cf.\ also 
\Cref{par:visualisation}).

\paragraph{Complexity.}  The training complexity is linear to the size of $\mathcal{T}$, which is the set of all triplets and bounded by $\mathcal{O}(n^3)$. However, a well chosen sampling strategy may decrease this bound.
It has been shown by \citep{jamieson2011low} that the minimum number of triplets to recover the ordinal embedding is $\Omega(nd\log n)$ in $\mathbb{R}^d$. We adapt this result to the setting in which the parameters to be learnt are a mean vector in $\mathbb{R}^d$ and a covariance matrix $\mathcal{S}^d_+$. Hence, the dimensionality can be considered to be $d'=d+d^2$ and $\mathcal{O}(d^2)$. Thus, the new recovered lower bound for the triplets becomes $\Omega(d^2n\log n)$, which is still polynomial in $d$ and $\mathcal{O}(n\log n)$. Since ordinal embeddings typically map into a low-dimensional space, this is not a drastic loss in efficiency.
Moreover, it is worth mentioning that a low number of epochs was needed for convergence for all experiments. 
Finally, the computational bottleneck when dealing with Wasserstein distance in its closed-form is computing the matrix square roots of the scale parameters. However, as we opted to learn diagonal covariances, hence this problem is not present in our approach.


\section{Experiments}
\label{sec:experiments}


\setlength\intextsep{0pt}
\begin{figure*}[!ht]
\centering
\subfigure[$\epsilon=0; p=\{1,2,4\}$]{%
\begin{tabular}{cc}
\includegraphics[width=1.75cm]{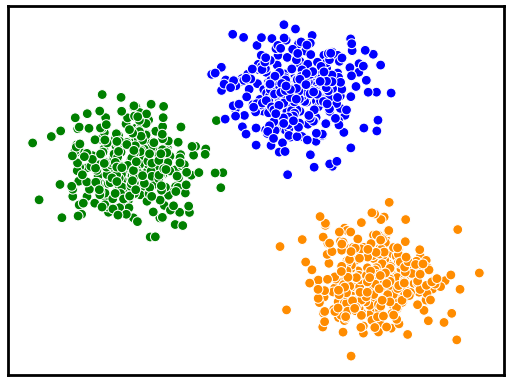}&
\includegraphics[width=1.75cm]{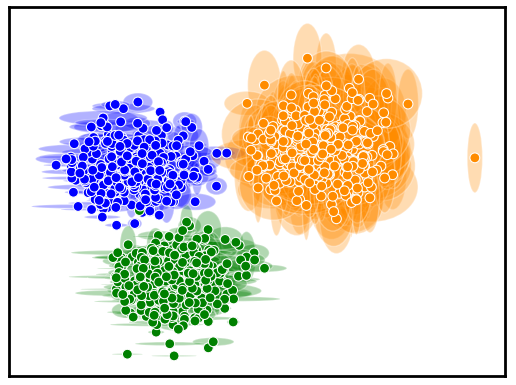}%
\includegraphics[width=1.75cm]{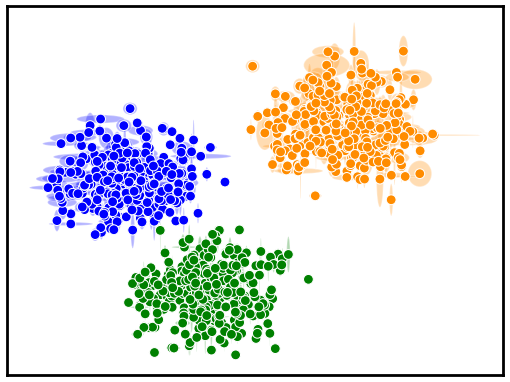}%
\includegraphics[width=1.75cm]{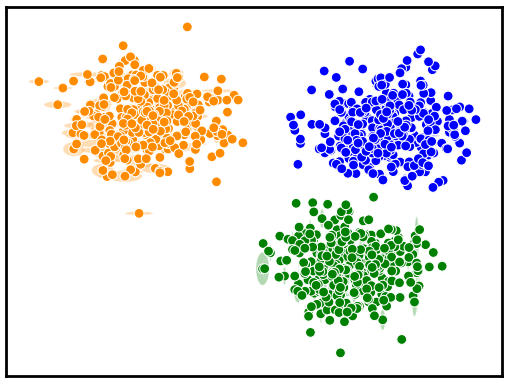}\\
\includegraphics[width=1.75cm]{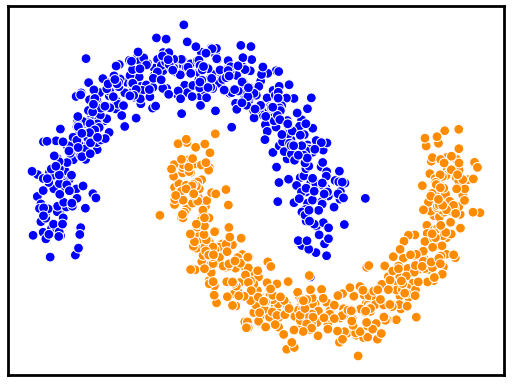}&
\includegraphics[width=1.75cm]{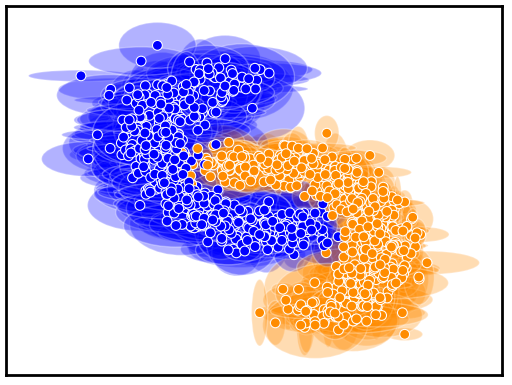}%
\includegraphics[width=1.75cm]{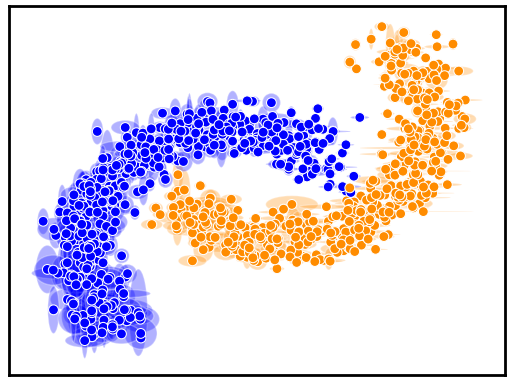}%
\includegraphics[width=1.75cm]{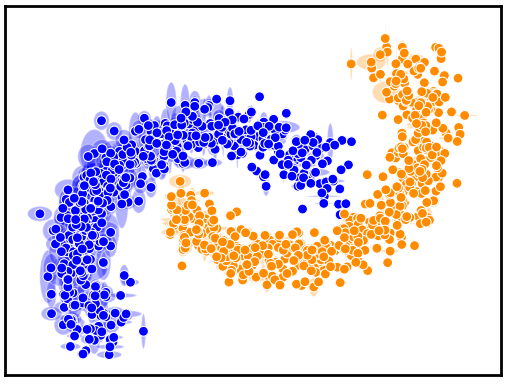}\\
\includegraphics[width=1.75cm]{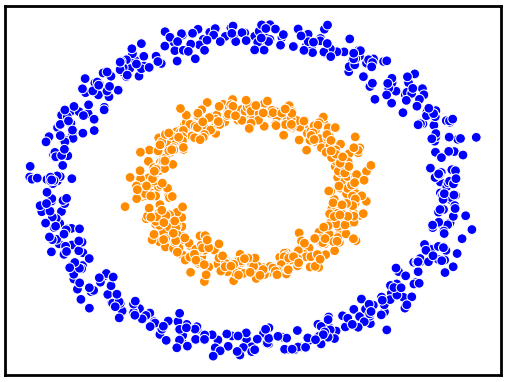}&
\includegraphics[width=1.75cm]{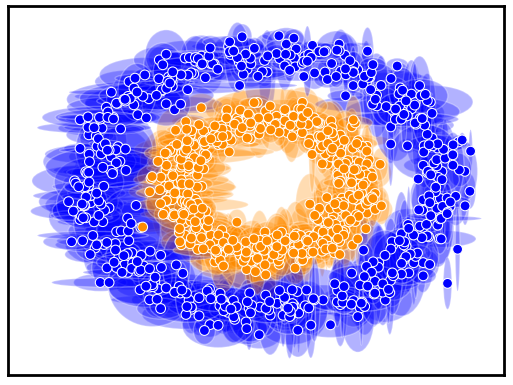}%
\includegraphics[width=1.75cm]{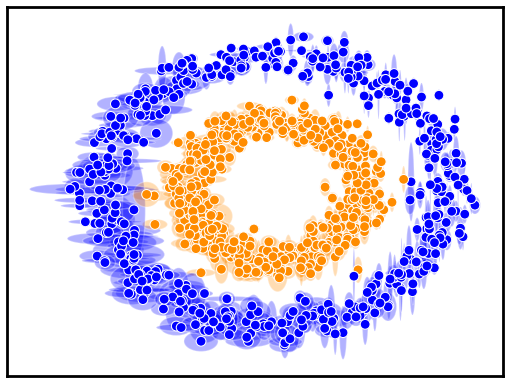}%
\includegraphics[width=1.75cm]{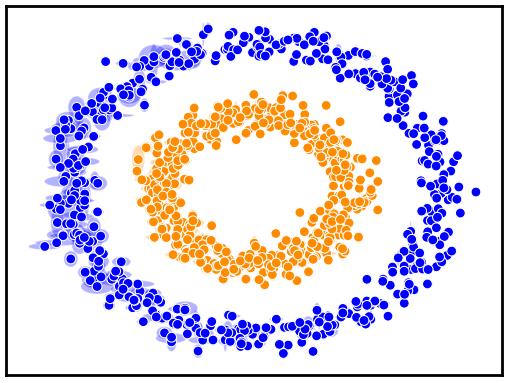}
\label{fig:toy_dataset_no_noise} 
\end{tabular}}
\hspace{1em}
\subfigure[$\epsilon=\{0.1, 0.2\}; p=4$]{%
\begin{tabular}{c}
\includegraphics[width=1.75cm]{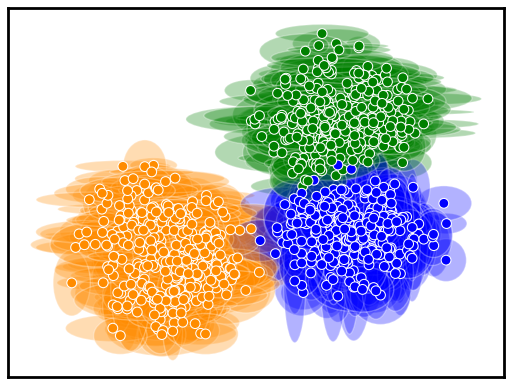}%
\includegraphics[width=1.75cm]{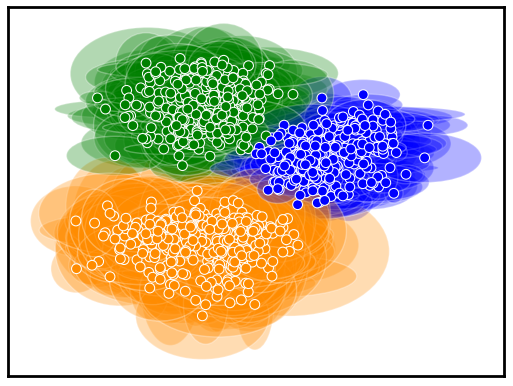}\\
\includegraphics[width=1.75cm]{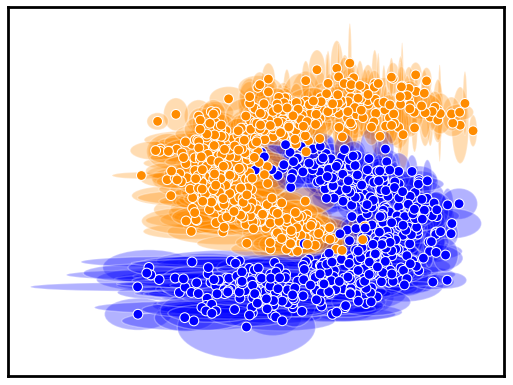}%
\includegraphics[width=1.75cm]{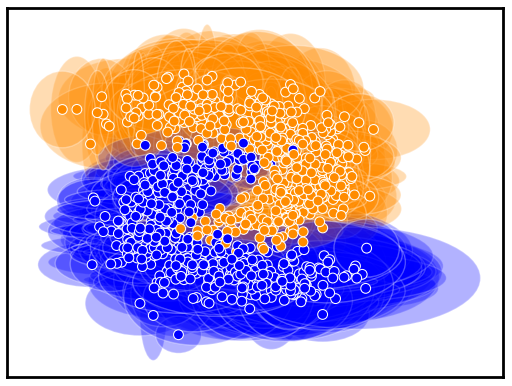}\\
\includegraphics[width=1.75cm]{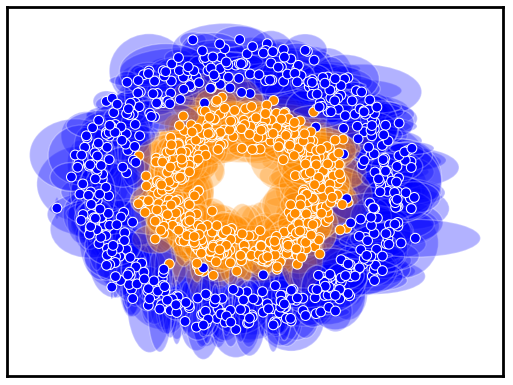}%
\includegraphics[width=1.75cm]{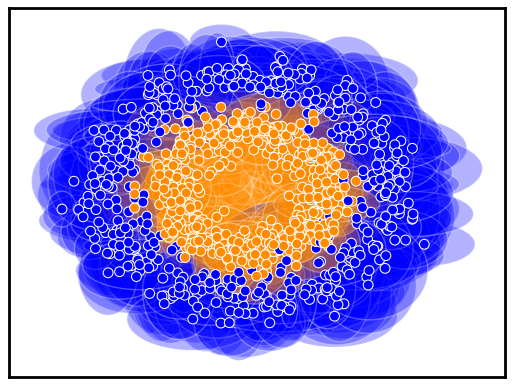}%
\label{fig:toy_dataset_noise} 
\end{tabular}}
\caption{\ElOE embeddings for synthetic experiments. From left to right, columns represent ground-truths, elliptical embeddings in nose-free setting ($\epsilon=0$) for increasing numbers of triplets $pd^2n \log n$ for $p=\{1,2,4\}$ (col. 2-4) and increasing noise for $p=4$ and $\epsilon=\{0.1,0.2\}$ (col. 5-6). Color indicates labels not used for training.} 
\end{figure*}

Our main objective is to investigate the effectiveness of elliptical embeddings for the problem of ordinal embedding. For this reason, we evaluate our method in two settings. First, we perform experiments on synthetic datasets in order to gain some insight regarding our approach. We then apply our approach to real datasets in order to assess the performance of our model in real cases.

As stated earlier, one common outcome of ordinal embedding methods is the reconstruction of dataset as well as density estimation. More specifically, in the ordinal embedding problem the distances can be recovered up to an orthogonal transformation. For this reason, when the ground truth is available, an adequate error metric to assess the embedding quality is the Procrustes distance \citep{dryden2016statistical}. 
We generalize the conventional definition to a distance between vectors and distributions as follows 
(the steps that lead to the generalization are provided in the appendix):

\begin{definition}[Procrustes Distance between distributions]
Given two finite sequences $X=(x_i)^n_{i=1}$, $X^{'}=(x^{'}_{i, \mu, \Sigma})^n_{i=1}$ in $\mathbb{R}^d$ of equal length with centroids in $\bar{x}$, $\bar{x}^{'}$ and centroids sizes $S_X$, $S_{X^{'}}$\footnote{Let us define the centroid $\bar{x}$ as $\bar{x} = \frac{1}{n} \sum^n_{i=1}x_i$, then the centroid size $S_X$ is 
$S_X = ( \frac{1}{n}\sum^n_{i=1}(\bar{x}-x_i)^2)^{1/2}$, provided we ignore the trivial case in which all points coincide.}, respectively, the Procrustes distance $d_P^\star(X,X^{'})$ between $X$ and $X^{'}$ is defined as:
\begin{equation}
    d_P^\star(X,X') = \inf_{R \in \mathcal{R} } \left( \sum^n_{i=1} \left\|\frac{Rx_i}{S_X} - \frac{\mu_{i}}{S_{X^{'}}} \right\|^2 + \frac{\Tr(\Sigma_i)}{S^2_{X^{'}}} \right)^\frac{1}{2}
    \label{eq:wasserstein_procrustes} 
\end{equation}
\end{definition}
\paragraph{Visualization of embeddings using ellipses}
\label{par:visualisation} The most significant difference between our distribution-based approach and the point-based embeddings is the variance. In particular, the variance has the purpose of reflecting the uncertainty. It does so by enriching the scope of the embeddings and by providing the possibility of continuously representing a discrete object in the metric space. 
In most cases related to multidimensional scaling, the output dimensionality of the representations is low, thus the learned embeddings can be visualized as they are. We argue that mapping objects into ellipses on a plane allows to better observe the relationship between objects visually. \citet{muzellec2018generalizing} state that visualizing the variances as they are is not natural to the human eye, and they instead favor a representation by the precision matrix rather than the covariance matrix. On the contrary, we believe that in this context a visualization based on the variance is preferable when the focus is on illustrating the spread around the location rather than the distance between the embeddings themselves.

\subsection{Reconstruction} 
\label{subsec:reconstruction}
We present empirical results that aim to evaluate the reconstruction abilities of the proposed approach. For this, we follow the same experimental setting of \citet{haghiri2019large}. More specifically, we use three 2-dimensional synthetic datasets generated with the \texttt{scikit-learn} package\footnote{https://scikit-learn.org/} in Python. The datasets are: a) two-moons dataset with two labels, b) the Blobs dataset, a mixture of three Gaussians $\mathcal{N}(\mu, \frac{1}{\sqrt{2}}I_2)$ and c) 2 concentric circles with 2 labels. For each dataset, $n=1000$ points are generated. The label information is used only for visualization purposes. 

We generate $|\mathcal{T}|$ random triplets sampled from a uniform distribution. To simulate the ordinal feedback from the oracle, we compute the difference of the squared $\ell_2$ norm between the points for a given triplet. The total number $|\mathcal{T}|$ is set to be $p d^2 n\text{log}n$ for $p=\{1,2,4 \}$. To evaluate the performance, we compute the triplet error as well as the Procrustes distance shown in \eqref{eq:wasserstein_procrustes}. 
\paragraph{Noise-free setting} 
In this series of experiments, we aim at investigating the influence of the number of triplets on the reconstruction ability, specifically on the variance of the elliptical embeddings. We first test in a noise-free setting. 
%
\Cref{fig:toy_dataset_no_noise} depicts the original datasets (to the left) and the learned embeddings for different values of $T$. From left to right, the number of used triplets $|\mathcal{T}|$ increases with $|\mathcal{T}|= p d^2 n\text{log}n$, where $p \in \{1,2,4\}$. 
For all three datasets, we observe that the reconstruction abilities w.r.t the location point improves when the number of triplets increases. Furthermore, we observe that on average the variance decreases with increasing $|\mathcal{T}|$, which confirms that the uncertainty about a point's location decreases when more exact comparisons are available. For example, for $p=4$, the average area of the ellipses is minimal. 
We can also observe that the variance enriches the visual representation. A point-vector representation may be misleading because when the algorithm is given few triplets, it has also to satisfy fewer constraints which means that the overall degree of freedom for selecting the individual points is greater. However 
a point-based visualization does not appreciate this fact. 
\paragraph{Noisy setting} Our next goal was to investigate the influence of noisy or erroneous triplets on the behaviour of the variance. We follow the 
procedure described above,
but simulated noise by randomly swapping the assessment of the oracle with a probability of  $\epsilon = \{0.1, 0.2\}$. 

\Cref{fig:toy_dataset_noise} shows the results obtained. %
We notice that when the proportion of erroneous triplets increases (from left to right), the variance on average increases for all triplets. Additionally, in order to quantitatively estimate the performance of our approach we measure the Procrustes distance $d_P^\star$ \Cref{eq:wasserstein_procrustes} with respect to the ground truths. As a baseline, we compare our model to \textsf{STE} (Stochastic Triplet Embedding) \citep{van2012stochastic}. For 10 rounds, we compute $d_P^\star$e \Cref{eq:wasserstein_procrustes} w.r.t $\epsilon$. We notice that generally, \ElOE  recovers better the density estimate even considering the variance of the ellipses.

\begin{figure}[ht]
\centering
\subfigure[Blobs dataset]{\label{fig:b_graph}\includegraphics[scale=0.2]{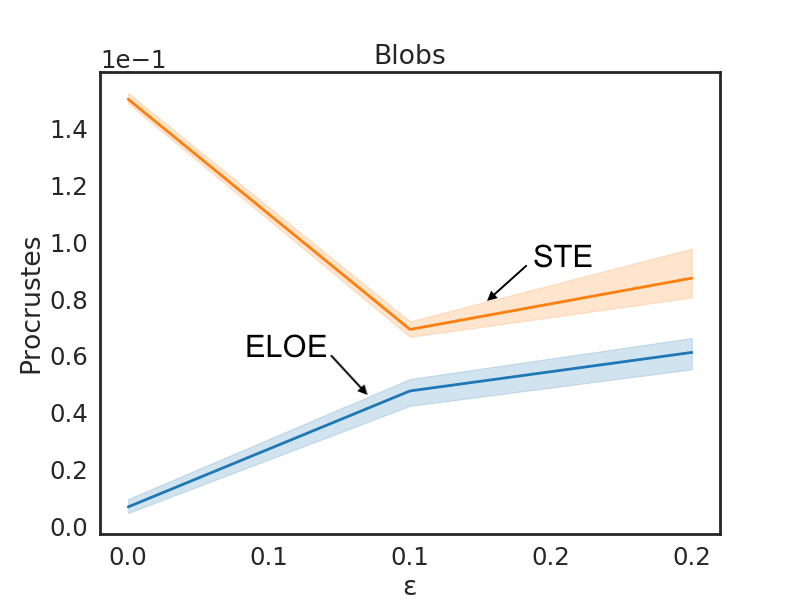}}
\hfill
\subfigure[Moons dataset]{\label{fig:m_graph}\includegraphics[scale=0.2]{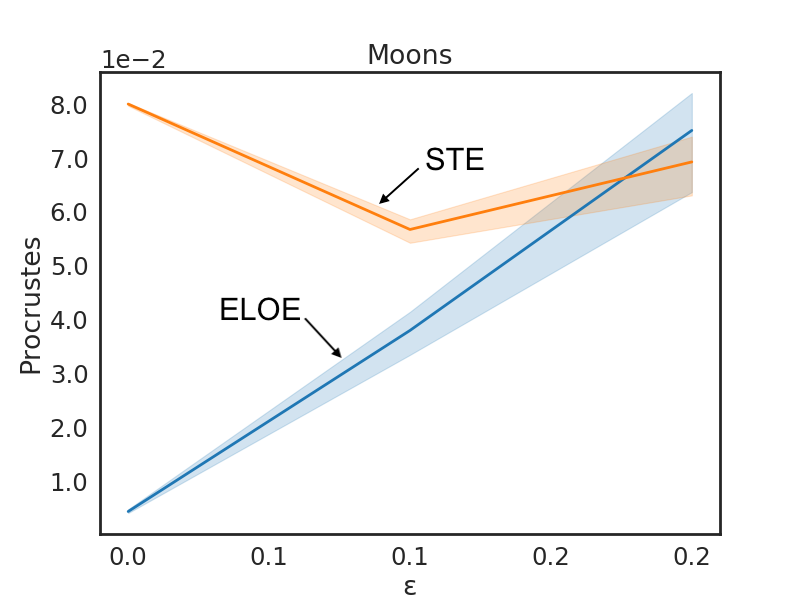}}
\hfill
\subfigure[Circles dataset]{\label{fig:c_graph}\includegraphics[scale=0.2]{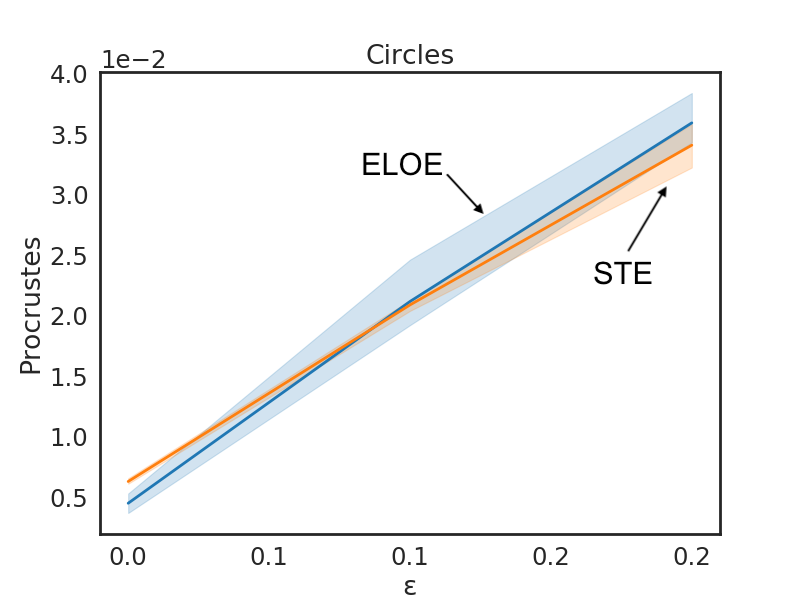}}
\hfill
\subfigure[Purity vs $n$]{\label{fig:mnist_clustering}\includegraphics[scale=0.2]{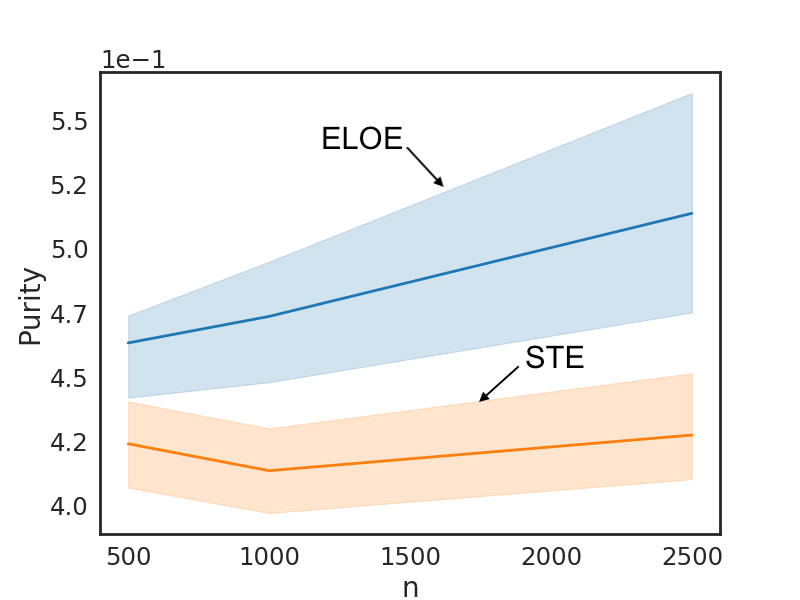}}
\caption{(a)--(c) Procrustes distance vs $\epsilon$ for synthetic datasets; (d) purity for MNIST dataset.}
\end{figure}

\subsection{Ordinal embedding} 
\paragraph{Food Dataset} We evaluate our method on the Food relative similarity dataset \citep{wilber2014cost}, humans were presented images of dishes and asked to compare similar dishes based on their taste. A good embedding method should show clusters of dishes of the same type. We compute two-dimensional embeddings of the food images based on the available unique triplets of 100 images with $|\mathcal{T}|=$190376.

\begin{wrapfigure}{r}{0.45\textwidth}
\includegraphics[scale=0.28]{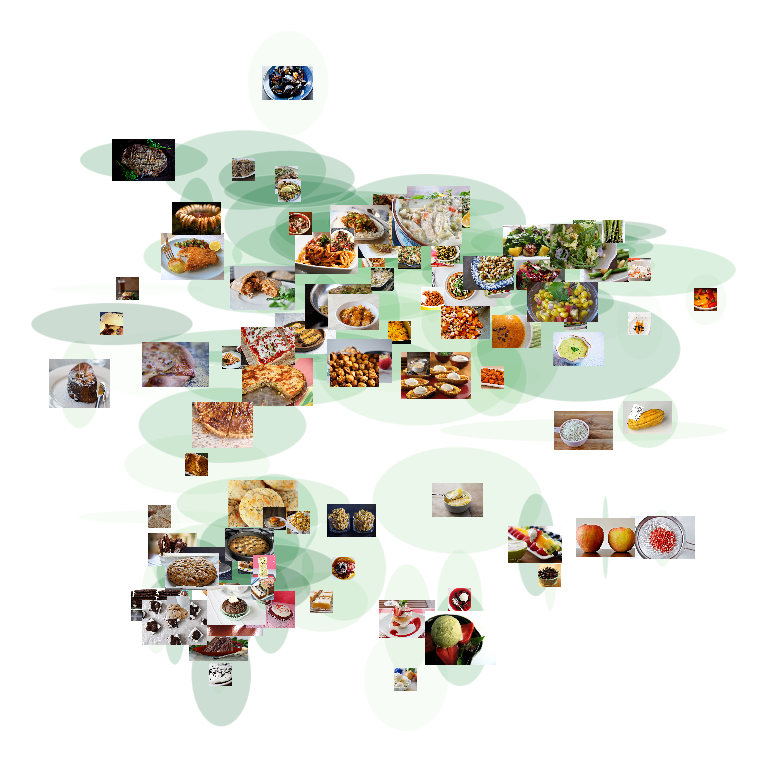}%
\label{fig:food_embeddings}
\caption{Elliptical embeddings for Food dataset.}
\label{fig:cifar_embedding}
\end{wrapfigure}

We compare our embeddings to \textsf{STE} and we observe that our embeddings in \Cref{fig:food_embeddings} closely match the one produced by \textsf{STE}, which is the reference model used by the authors of the dataset. A full scale image is provided in the appendix as well as dendrograms of clusters. This dataset has also 9349 pairs of contradicting triplets. Note that for this dataset, no ground truth is available and hence there is no way other than visual inspection for evaluating our results. 


\paragraph{MNIST Dataset} 
On this dataset, we reproduce the experiment conducted in \citep{kleindessner2017kernel}. For $n=500, 1000$ and $2500$, we uniformly chose $n$ MNIST digits randomly and we generate $200n\log n$ triplets comparisons based on the Euclidean distances between the digits. Each comparison is incorrect with probability $\epsilon=0.15$. We then generate an ordinal embedding with $d=5$ and compute a k-means clustering on the obtained embeddings. 
\Cref{fig:mnist_clustering} compares the purity of the clusters obtained with \textsf{STE} and \ElOE embeddings. Purity is computed as $\text{purity}(\Omega, \mathcal{C})=n^{-1}\sum_k |w_k \cap c_j|$, where the clusters are $\Omega=\{w_1, \ldots, w_k\}$ and the classes are $\mathcal{C}=\{c_1, \ldots,c_j\}$. High purity is better.
In order to take into account the variance, we concatenate the diagonal of the covariance matrix and the mean vector for each embedding. We observe that the purity of the clustering from \ElOE is consistently higher for all values of $n$ considered.
\begin{figure}[ht]
\centering
\subfigure[$\epsilon = 0$]{\label{fig:mnist_lin_ord_0}\includegraphics[scale=0.1]{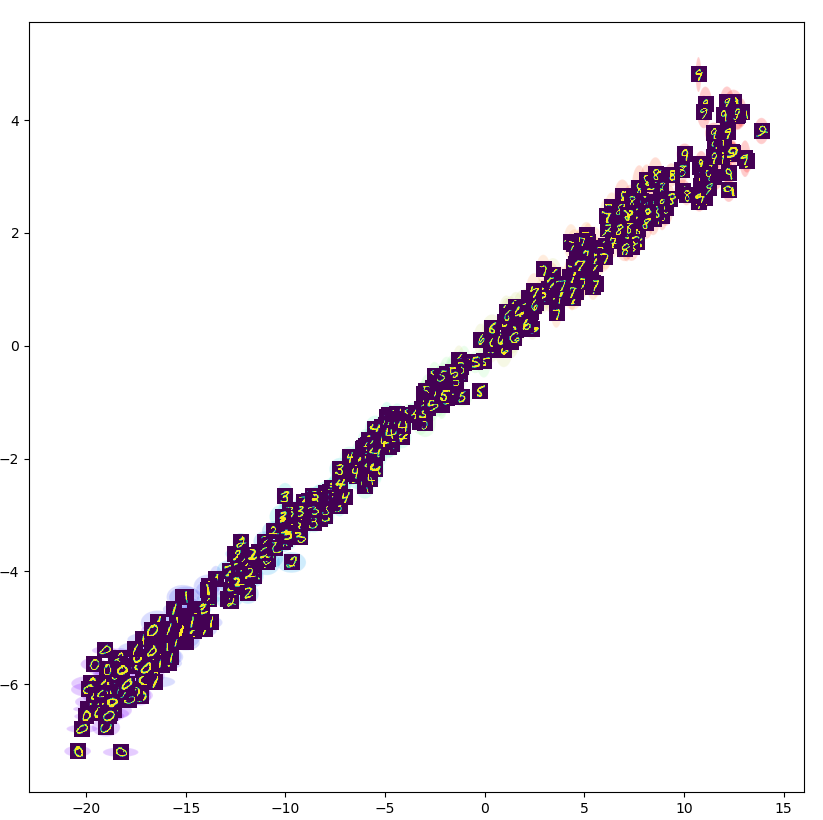}}
\hfill
\subfigure[$\epsilon=0.1$ ]{\label{fig:mnist_lin_ord_10}\includegraphics[scale=0.1]{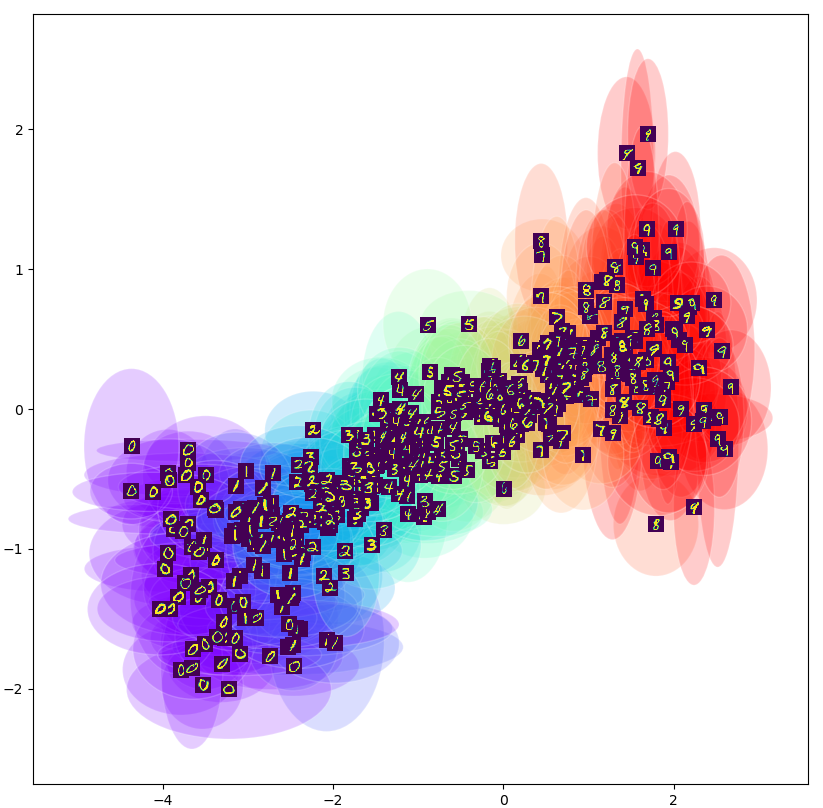}}
\hfill
\subfigure[ $\epsilon=0.2$]{\label{fig:mnist_lin_ord_20}\includegraphics[scale=0.1]{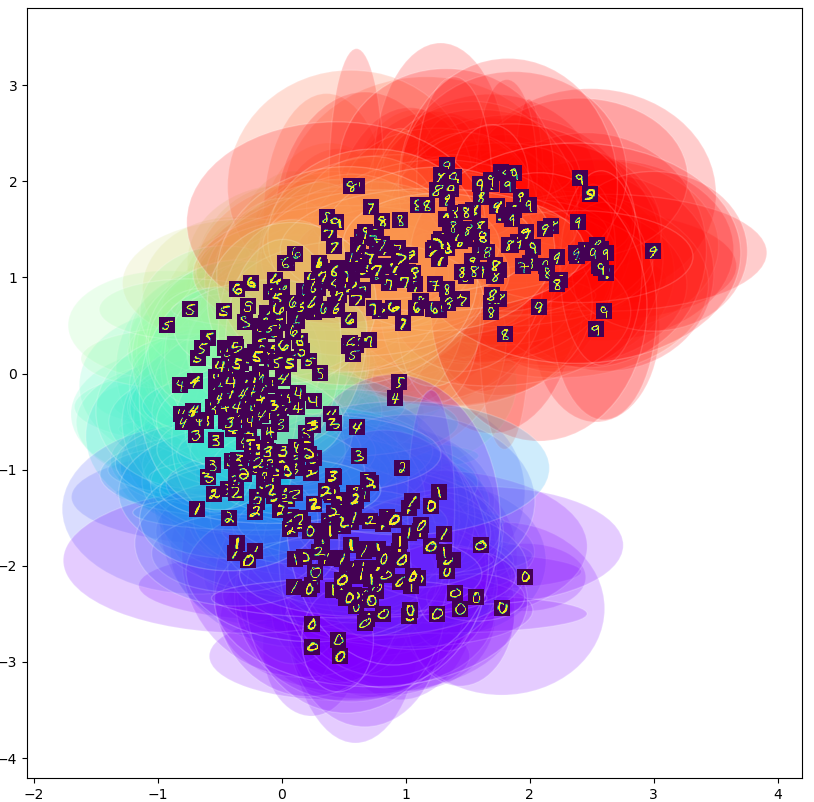}}
\hfill
\subfigure[ $\epsilon=0.3$]{\label{fig:mnist_lin_ord_30}\includegraphics[scale=0.1]{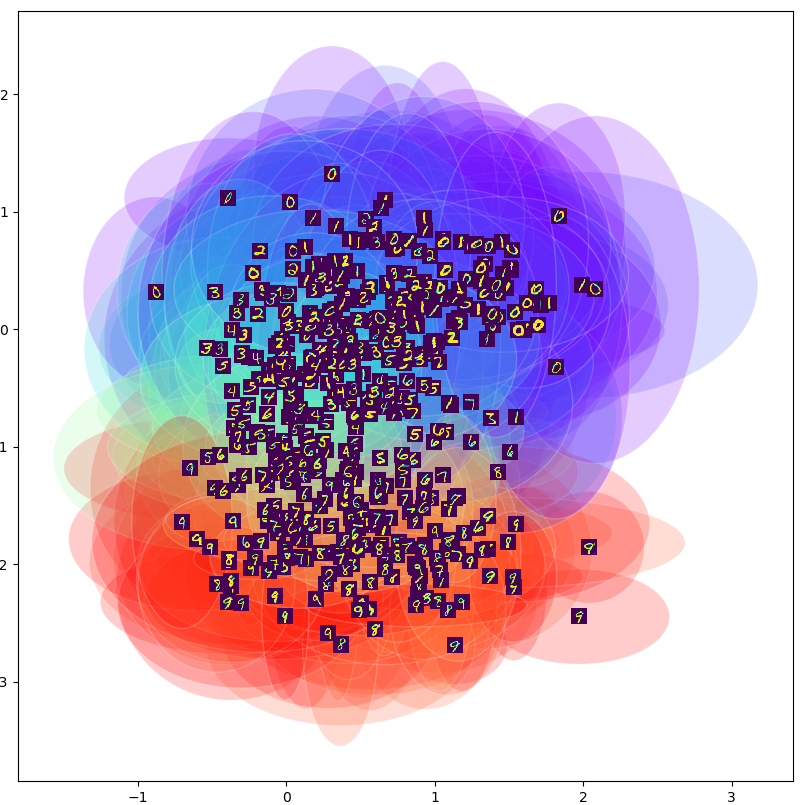}}
\caption{Linear order in the MNIST dataset. Color indicates the label of the handwritten digits to better appreciate the linearity.}
\label{fig:mnist_linear_order}
\end{figure}

\subsection{Semantic Embedding}
In this section, we intend to embed a real-world full or partial ordinal relation between data points, in this case images. In particular we study the following three types of relations, where the given ordinal relation is derived from various label structures of the objects, such as a linear or a hierarchical order.
Intuitively, we want all nodes that belong to the 1-hop neighborhood of item $i$ to be closer to $i$ in their embedding, compared to the nodes in the 2-hop neighborhood, which in turn will be closer than the items in the 3-hop neighborhood and so on. 
Moreover, we need to adapt the sampling strategy to deal with this task because uniformly sampling triplets leads to oversampling more frequent high-degree nodes. 
Thus, we use the following strategy: we first sample a node $i$, then we sample a node from each of its neighborhoods, and randomly choose one of those triplets. 

\begin{wrapfigure}[33]{r}{0.3\textwidth}
\subfigure[CIFAR]{\label{fig:cifar_embedding}\includegraphics[scale=0.2]{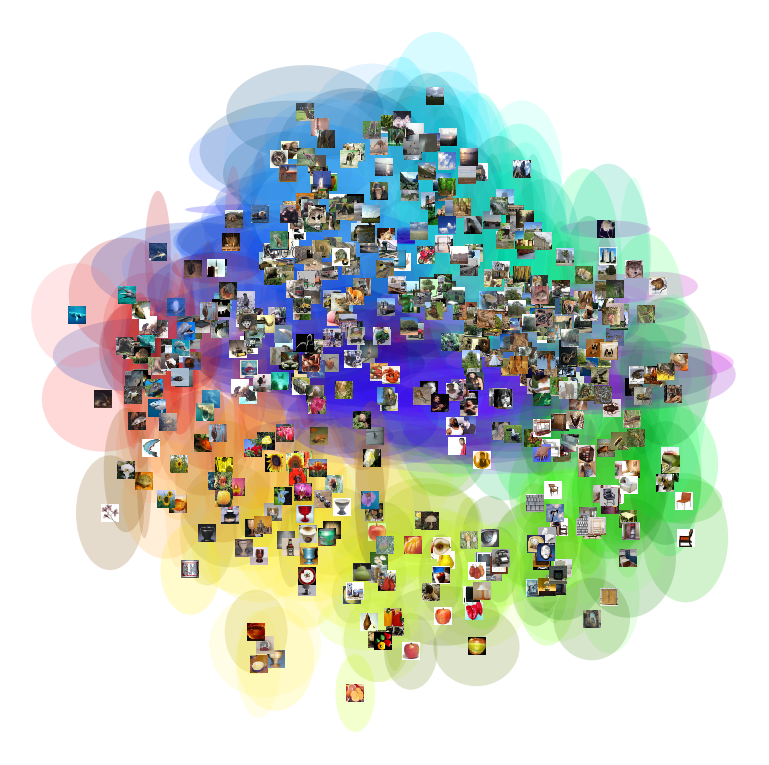}}
\hfill
\subfigure[VOC]{\label{fig:voc_embedding}\includegraphics[scale=0.2]{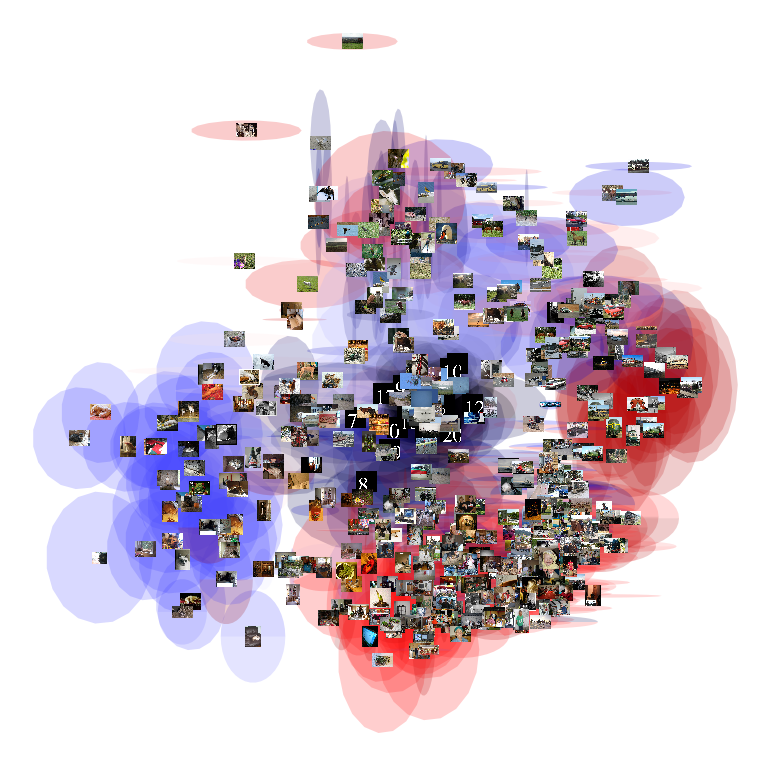}}
\caption{Visualization of the embeddings for CIFAR And VOC.}
\label{fig:mnist_linear_order}
\end{wrapfigure}

\paragraph{Linear order}
This simple experiment aims at verifying whether our approach is able to capture the structural information when the underlying ordinal feedback derived from labels is a linear order with MNIST dataset. In this case, we use the same sampling strategy described in  \Cref{subsec:reconstruction}. 
We train our model with $200n\log n$ triplets and we sample 500 digits for visualization. We perturbed a subset of the available triplets defined by $\epsilon=\{0, 0.1, 0.2, 0.3\}$. \Cref{fig:mnist_linear_order} illustrates the  obtained embeddings.

In the noise-free case (\Cref{fig:mnist_lin_ord_0}) we obtain a linear relation in which the embedding have very small variances. When perturbing $10\%$ of the triplets, as shown in \Cref{fig:mnist_lin_ord_10} the linear order is maintained but we can observe an average increase in the variance of the embeddings. Finally, for $\epsilon = 0.3$ (\Cref{fig:mnist_lin_ord_30}), the proportion of noisy triplets is so high that even the linear order is perturbed. Nevertheless, the clusters defined by the classes of the embeddings are still easily identified. Additional numerical results are presented in the supplemental material section.

\paragraph{Hierarchical relation}
This experiment was conducted on CIFAR100, a multi-class image dataset 
where each of the 60000 images has two different levels of labels, a super-class label and a fine-class label. 
The semantics of the dataset is such that there are 20 super-classes,
each of which has 5 labels. The graph structures can be seen in the appendix. We sample $n=5000$ images to create $2 nd^2\log n$ triplets. The triplet score is computed with the methodology described earlier, through the shortest path distance between nodes.
The result of the embeddings is shown in \Cref{fig:cifar_embedding} for 500 randomly sampled images. We can notice how the structural information of the graph is retrieved. In fact, we see that images, fine-classes and super-classes form concentric circles, the images being the most exterior.

\begin{wrapfigure}[8]{r}{0.3\textwidth}
\scalebox{0.85}{
\begin{tabular}{l cc cc}
& \multicolumn{2}{c}{CIFAR} & \multicolumn{2}{c}{VOC} \\
\cmidrule(lr){2-3} \cmidrule(lr){3-5}
     & AUC   & AP   & AUC  & AP \\
\midrule
$\textsf{STE}$  & 0.47 & 0.53 & 0.54  & 0.56 \\
Baseline & 0.88 & 0.90 & 0.93  & 0.94 \\
$\textsf{ELOE}$ & 0.89 & 0.92 & 0.95  & 0.95 \\
\bottomrule
\end{tabular}}
\caption{Link prediction scores.}
\label{tab:link_prediction}
\end{wrapfigure}

To quantitatively assess the meaningfulness of the embeddings, we report the area under the ROC curve (AUC) and the average precision (AP) of randomly sampled triplets. 
We  compare our results to \citep{haghiri2019large} which we re-implemented with the same hidden size of \ElOE and \textsf{STE}. It is worth noticing that this method can also be seen as the producing distributional vectors with null variance.
The score considered for \ElOE is $E_{ij}$, $\|\mathbf{x}_i - \mathbf{x}_j \|^2$ for \textsf{STE} and \citep{haghiri2019large}, where $\mathbf{x}_{i}$ is the embedding of item $i$. Results are reported in \Cref{tab:link_prediction}. We see that \ElOE embeddings satisfy more triplets than \textsf{STE} embeddings. 

\paragraph{Multilabel distance}
Finally, we looked at the PASCAL VOC multi-label dataset, where each image can be assigned to multiple labels. Here $n=5000$ and $p=2$. 
The same concerns with respect to the sampling strategy occur in this case as well, and we apply the same methodology described earlier. In this case, an image node can be connected to multiple node classes. The obtained results showed in \Cref{fig:voc_embedding} confirm our intuition, nodes with less diverse neighborhoods have a lower variance, hence less uncertainty compared to nodes that belongs to multiple classes. In fact, the inclusion in multiple classes makes the embedding location less certain.  


\section{Conclusion}

We have proposed to generalize the ordinal embedding problem by mapping objects in the space of Gaussian distributions endowed with the Wasserstein distance. This is based on the generalization of point embeddings in $\mathbb{R}^d$ to distributions. Each embedding is described by a location parameter $\mu$ and a scale parameter $\Sigma$, visualized as ellipses. We argue that this allows to more informative perceptual embeddings by representing uncertainty of the representation. 
In a number of experiments on different datasets we demonstrate the validity of our approach. We show that the proposed framework is robust and beneficial when the triplet comparisons are noisy.  Overall, with our proposed approach we are able to obtain valid embedding that can be used for downstream tasks. As future work we aim to study other distributions beyond Gaussian for the problem of ordinal embedding.

\bibliographystyle{unsrtnat}
\bibliography{ref}

\appendix

\section{Appendix}
\subsection{Procrustes Distance}
The Procrustes distance is defined as follows:

\begin{definition}[Procrustes Distance]

Given two finite sequences $X=(x_i)^n_{i=1}$, $X^{'}=(x_{i}^{'})^n_{i=1}$ in $\mathbb{R}^d$ of equal length with centroids in $\bar{x}$, $\bar{x}^{'}$ and centroids sizes $S_X$, $S_{X^{'}}$ \footnote{Let us define the centroid $\bar{x}$ as $\bar{x} = \frac{1}{n} \sum^n_{i=1}x_i$, then the centroid size $S_X$ is 
$S_X = ( \frac{1}{n}\sum^n_{i=1}(\bar{x}-x_i)^2)^{1/2} $}, respectively, the Procrustes distance $d_{P}(X,X^{'})$ between $X$ and $X^{'}$ is defined by:

\begin{equation}
    d_P(X,X') = \inf_{R \in \mathcal{R} } \left( \sum^n_{i=1} \left\| \frac{Rx_i}{S_X} - \frac{x_i^{'}}{S_{X^{'}}} \right\|^2\right)^{1/2}
\label{eq:procrustes_classic}
\end{equation}
where $\mathcal{R}$ is the group of Euclidean transformations (reflections, rotations and translations).
\end{definition}

We propose the following generalization: 

\begin{definition}[Procrustes Distance between vectors and distributions]
Given two finite sequences $X=(x_i)^n_{i=1}$, $X^{'}=(x^{'}_{i, \mu, \Sigma})^n_{i=1}$ in $\mathbb{R}^d$ of equal length with centroids in $\bar{x}$, $\bar{x}^{'}$ and centroids sizes $S_X$, $S_{X^{'}}$, respectively, the Procrustes distance $d_{P}(X,X^{'})$ between $X$ and $X^{'}$ is defined by:
\begin{equation}
    d_P^\star(X,X') = \inf_{R \in \mathcal{R} } \left( \sum^n_{i=1} \left\|\frac{Rx_i}{S_X} - \frac{\mu_{i}}{S_{X^{'}}} \right\|^2 + \frac{\Tr(\Sigma_i)}{S^2_{X^{'}}} \right)^{1/2}
\label{eq:procrustes_wassertein}
\end{equation}
\end{definition}


Let us rewrite the Procrustes distance from \cref{eq:procrustes_classic} as, where $\delta$ is a distance measure:
\begin{flalign*}
d_P(X,X^\prime) & = \inf_{R \in \mathcal{R} } \left( \sum^n_{i=1} \delta \left( \frac{Rx_i}{S_X} - \frac{x_i^\prime}{S_{X^\prime}} \right)^2\right)^{1/2} & \\
x_i^\prime & \sim  \mathcal{N}(\mu_i, \Sigma_i) \\
\frac{x_i^\prime}{S_{X^\prime}} & \sim  \mathcal{N}\left(\frac{\mu_i}{S_{X\prime}}, \frac{\Sigma_i}{S^2_{X^\prime}}\right)
\end{flalign*}

Let us define $\delta$ as the 2-Wasserstein distance $W_2$. Then: 
\begin{flalign*}
d_P(X,X\prime) & = \inf_{R \in \mathcal{R} } \left( \sum^n_{i=1} W_2 \left( \frac{Rx_i}{S_X},\frac{x_i^\prime}{S_{X^\prime}} \right)^2\right)^{1/2} & \\
 & \forall i \in |X^\prime|, \; \forall S_{X^\prime} \in \mathbb{R}_{>0}
\end{flalign*}

It has been established that the Wasserstein distance between a vector $\mathbf{h}$ and a Gaussian distribution $\nu \sim \mathcal{N}(\mu_\nu, \Sigma_\nu)$ is \citet{muzellec2018generalizing}:
\begin{flalign*}
W^2_2(\mathbf{h}, \nu ) & = \|\mathbf{h} - \mathbf{\mu}_\nu \|^2 + \Tr(\Sigma_\nu) & 
\end{flalign*}
Hence: 
\begin{flalign*}
d_P(X,X^\prime) & = \inf_{R \in \mathcal{R} } \left( \sum^n_{i=1} \left\| \frac{Rx_i}{S_X} - \frac{\mu_i}{S_{X^\prime}} \right\|^2 + \Tr\left(\frac{\Sigma_i}{S^2_{X^\prime}}\right) \right)^{1/2} &
\end{flalign*}
Using the linearity of the trace operator:
\begin{flalign*}
d_P(X,X^\prime) & = \inf_{R \in \mathcal{R} } \left( \sum^n_{i=1} \left\| \frac{Rx_i}{S_X} - \frac{\mu_i}{S_{X^\prime}} \right\|^2 + \frac{\Tr(\Sigma_i)}{S^2_{X^\prime}}  \right)^{1/2} &
\end{flalign*}

\subsection{Architecture and hyperparameters}
We noticed that \textsf{ElOE} is not very sensitive to the choice of number and size of hidden layers. For such, we chose a sufficiently large hidden size, specifically $h_{dim}=50$. To obtain the embeddings for an item indexed $i$ we have:
\begin{gather*}
\mathbf{h}_i=\text{relu}(\mathbf{x}_i\mathbf{W}_i + \mathbf{b}) \\
\mu_i=\mathbf{h}_i\mathbf{W}_{\mu} + \mathbf{b}_{\mu} \\
\sigma_i=\text{exp}(\mathbf{h}_i\mathbf{W}_{\Sigma} + \mathbf{b}_{\Sigma})
\end{gather*}

where $\mathbf{x}_i$ is a random sample from $\mathcal{N}(0,I_{h})$ and relu is the rectifier linear unit. We apply the exponential function to make sure that $\sigma_i$ is positive (and $\Sigma_i$ is positive definite).
Weight matrices $\mathbf{W}_\mu$, $\mathbf{W}_\Sigma$ and $\mathbf{W}$ are initialized with Xavier initialization 
. As stated earlier, we do not regularize the norm of the mean vectors but we bound the values of the covariance matrices with $C=\log(100)$. However, we observe that this additional precaution is not needed unless the number of contradicting triplets is too large. This is due to the self-regularizing nature of the Wasserstein distance and it was confirmed by our experiments in which the average value of the variance is far from that bound for reasonable levels of noise.
All parameters are optimized using Adam 
, with a fixed learning rate of 0.01 and a learning rate decay of $10^-5$.

\subsection{Ethical considerations}

There is an inherent risk of incorporating implicit biases from data for any application that learns directly from the data itself.
Our work aims at mapping complex data in a low-dimensional space expressing the uncertainty and noise of the data itself and it relies on comparisons produced by human workers. In fact, it is well known that ordinal embedding was invented as a tool of visualization for psychometric data 
. However, producing robust visualizations such the ones our model proposes could identify these biases and spotlighting them. For these reasons, we state that this work does not present any foreseeable societal consequence but rather it could be use as a tool to analyse data for which bias could be difficult to foresee.

\subsection{Additional numerical results}
\Cref{tab:te_mnist} presents the empirical error for corresponding to \cref{fig:mnist_linear_order}. The values were obtained after averaging 10 runs for different values of $\epsilon=\{0, 0.1, 0.2, 0.3\}$. In order to better compare the values, we consider the triplet error when training point vectors as in \citet{haghiri2019large} and the triplet error for distributional embedding from $\textsf{ELOE}_{(\mu,\Sigma)}$ but only considering the center of the ellipses. We obtained very similar values, which prove that we indeed obtain plausible and correct embeddings. 
\vspace{.5cm}
\begin{table}[h]
\centering
\begin{tabular}{l c c c c} 
\toprule
& $\epsilon=0$ & $\epsilon=0.1$ & $\epsilon=0.2$ & $\epsilon=0.3$\\ 
\midrule
\cite{haghiri2019large} & 0.00 & 0.01 & 0.03 & 0.08 \\ 
$\textsf{ELOE}_{(\mu,\Sigma)}$ & 0.00 & 0.01 & 0.03 & 0.09 \\
\bottomrule
\end{tabular}
\caption{Triplet error results for the MNIST dataset.}
\label{tab:te_mnist}
\end{table}

\subsection{Omitted figures}
\subsubsection{Food Dataset embedding visualization}
\begin{figure*}[h!]
\centering
\subfigure[]{\label{fig:ste_food}\includegraphics[scale=0.5]{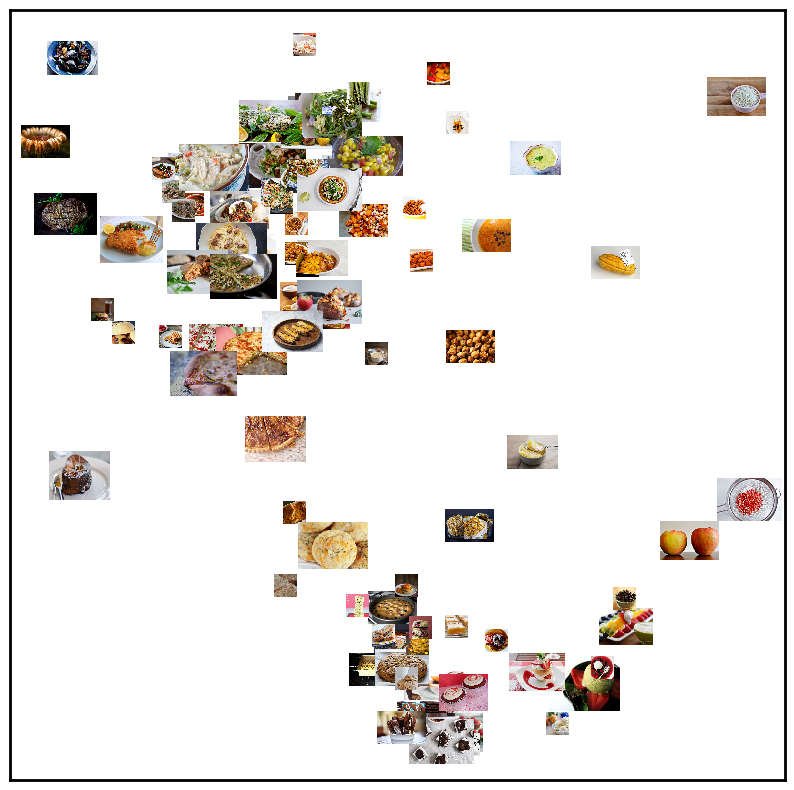}}
\subfigure[]{\label{fig:eloe_food}\includegraphics[scale=0.5]{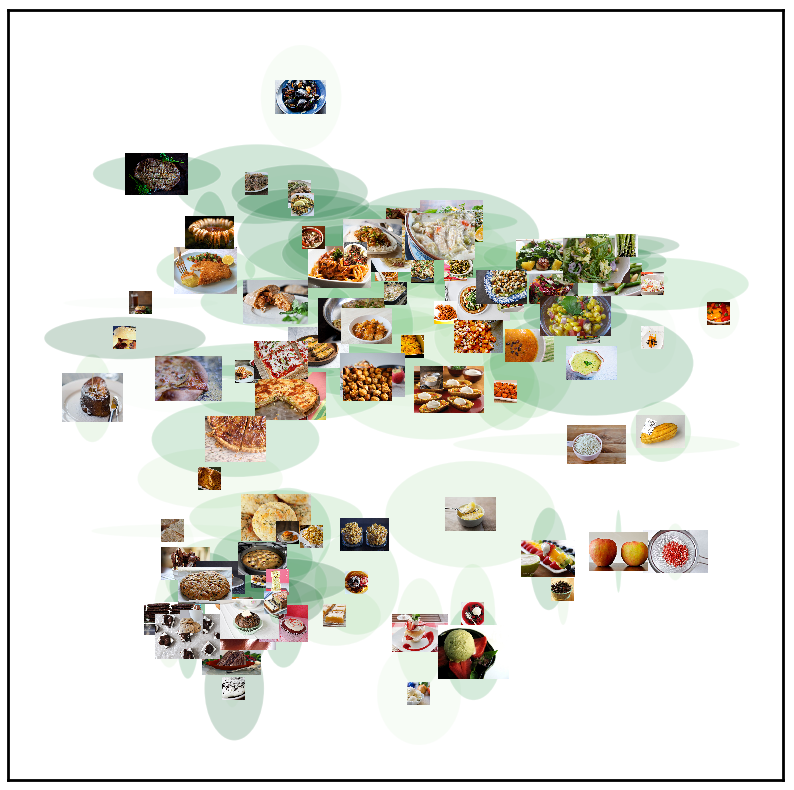}}
\caption{Visualization of the food dataset embeddings. (a) \textsf{STE}, (b) \textsf{ElOE}. }
\label{fig:food_big}
\end{figure*}

In this section, we present additional figures for better visualizing the Food dataset embeddings. \cref{fig:eloe_food_dendrogram} and \cref{fig:ste_food_dendrogram} depict the dendrograms of the hierarchical clustering for \textsf{STE} and \textsf{ElOE} embeddings, respectively. Closely analyzing each of the ten clusters, we can notice that they are quite homogenous w.r.t the taste of the dishes ( e.g all desserts being grouped together or the salads). \cref{fig:food_big} compares the embedding generated from \textsf{STE} and \textsf{ElOE}.

\begin{figure}[!h]
\centering
\subfigure[]{\label{fig:ste_food_dendrogram}\includegraphics[scale=0.41]{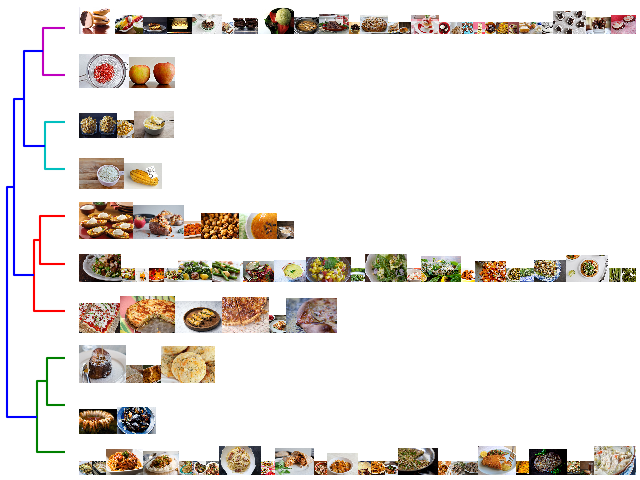}}
\hfill
\subfigure[]{\label{fig:eloe_food_dendrogram}\includegraphics[scale=0.43]{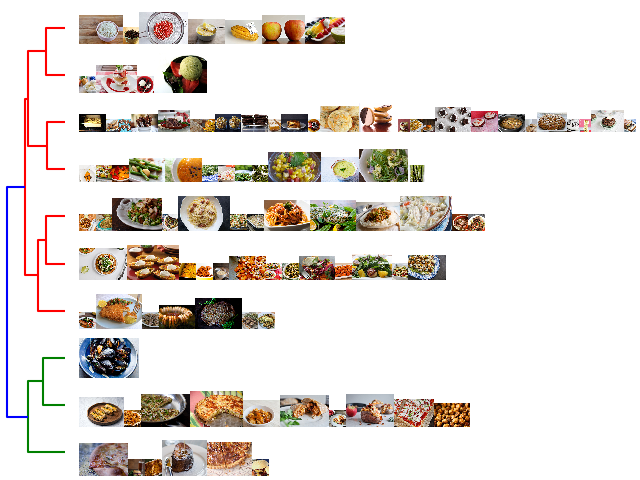}}
\caption{Dendrogram of the clustering  for the Food dataset. (a) \textsf{STE}, (b) \textsf{ElOE}. }
\end{figure}

\begin{figure}[!h]
\centering
\subfigure[]{\label{fig:ste_food_dendrogram}\includegraphics[scale=0.2]{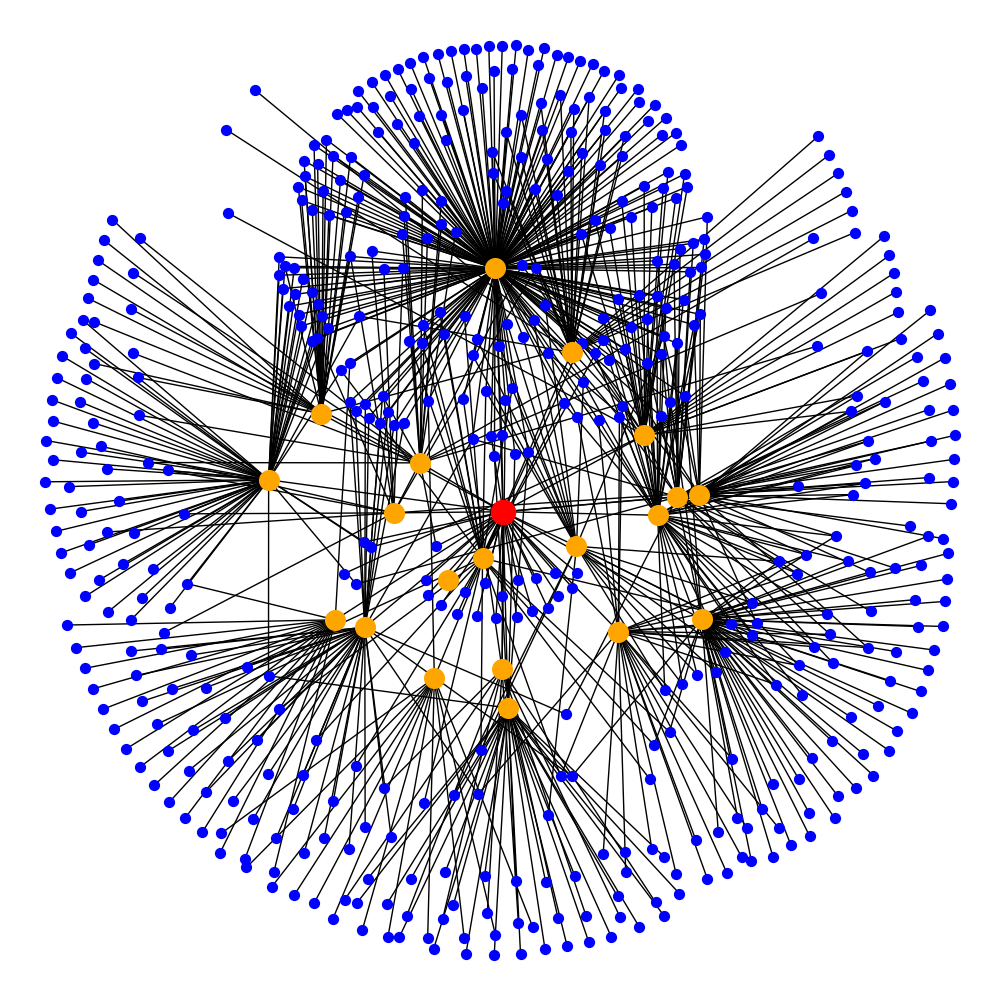}}
\hfill
\subfigure[]{\label{fig:eloe_food_dendrogram}\includegraphics[scale=0.2]{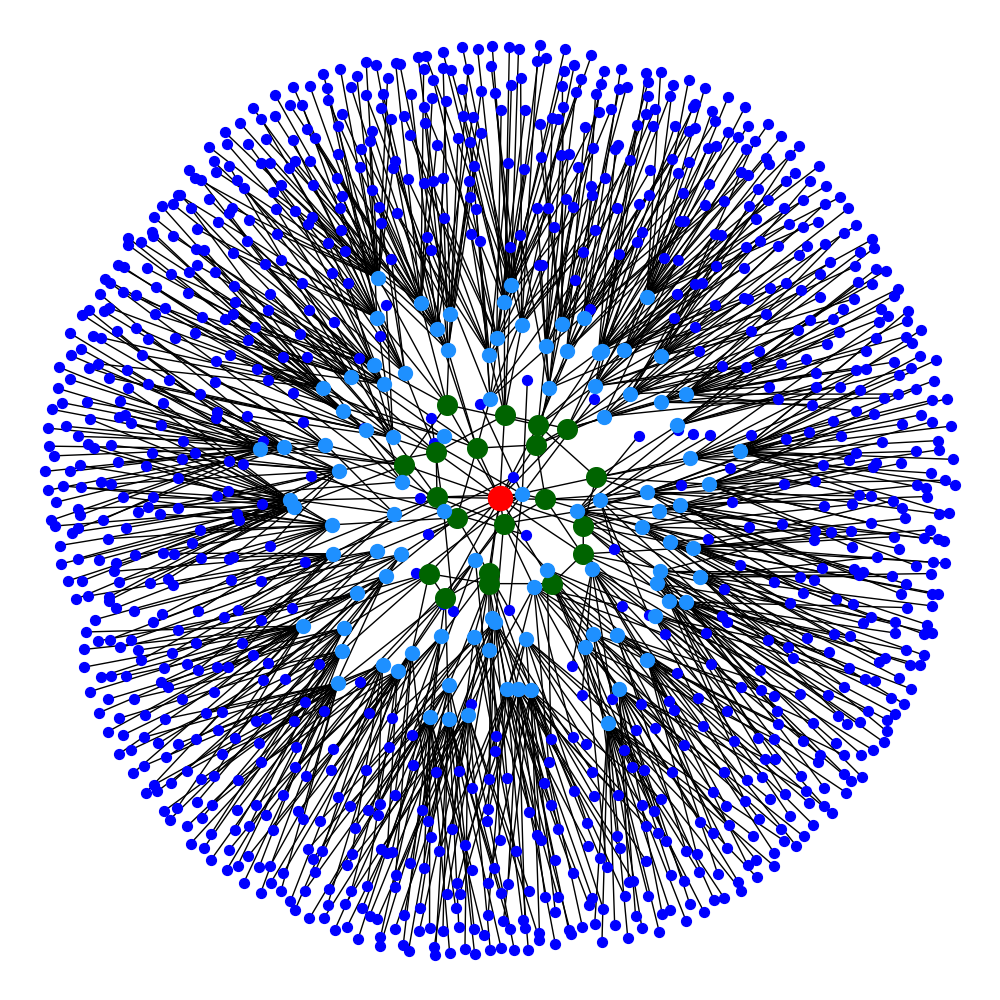}}
\caption{Graph of dataset for the semantic embedding task. Blue dots are image nodes, cyan dots are fine-class nodes and green dots are super-class nodes (a) {VOC}, (b) {CIFAR-100}. }
\end{figure}

\subsubsection{Synthetic dataset}
\begin{figure*}[ht!]
\centering
\begin{tabular}{ccc}
Ground-truth & Distance matrix & Similarity scores \\
\includegraphics[width=2.5cm]{images/blobs.png}&
\includegraphics[width=2.cm]{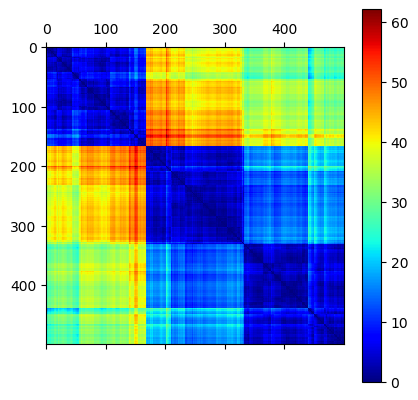}%
\includegraphics[width=2.cm]{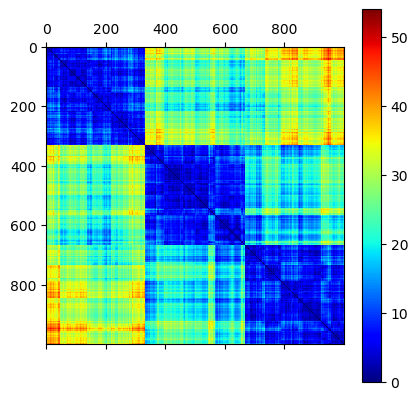} &
\includegraphics[width=2.5cm]{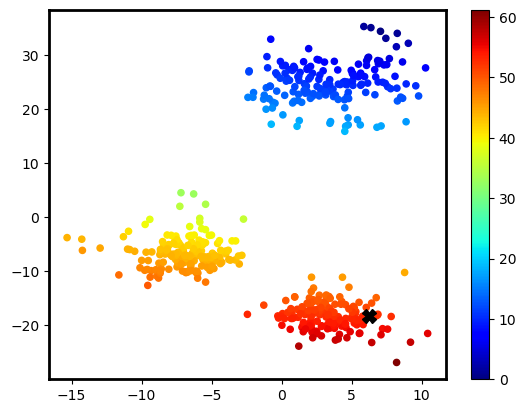}%
\includegraphics[width=2.5cm]{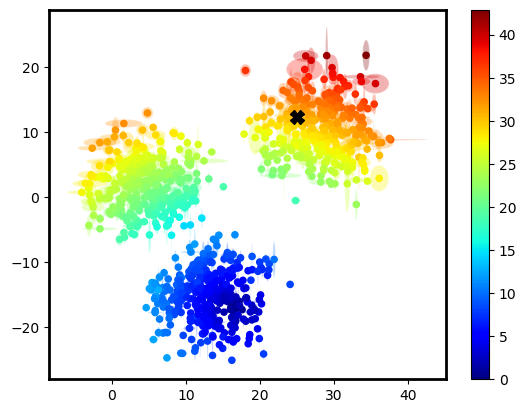} \\
\includegraphics[width=2.5cm]{images/moons.png}&
\includegraphics[width=2.cm]{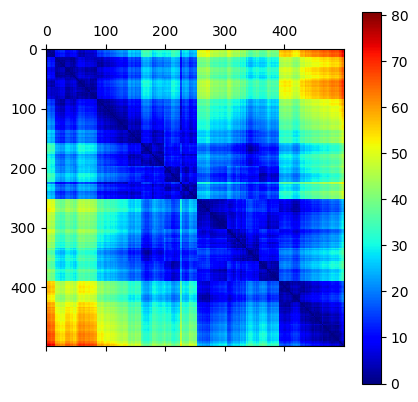}%
\includegraphics[width=2.cm]{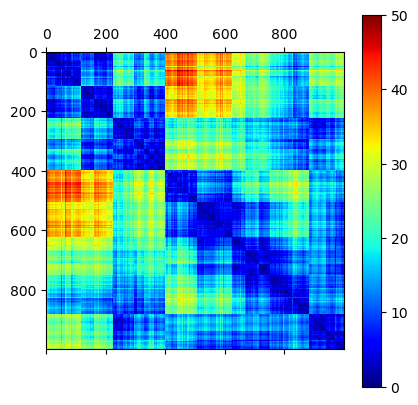} &
\includegraphics[width=2.5cm]{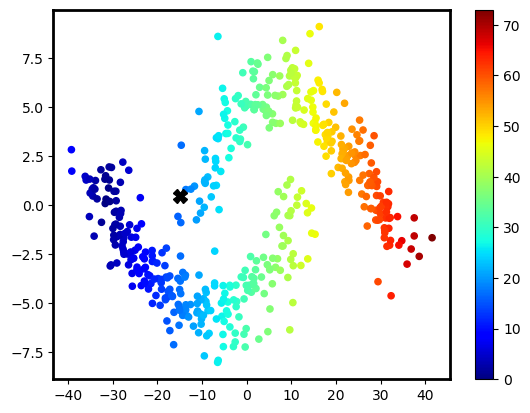}%
\includegraphics[width=2.5cm]{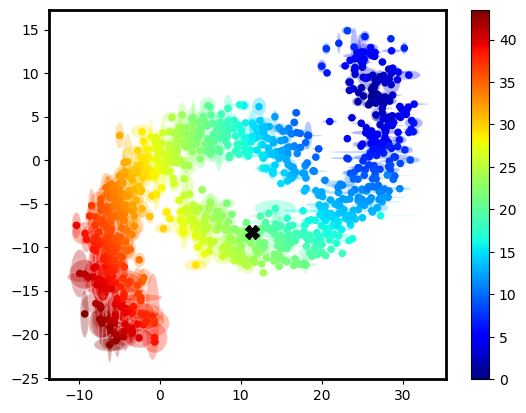} \\
\includegraphics[width=2.5cm]{images/circles.png}&
\includegraphics[width=2.cm]{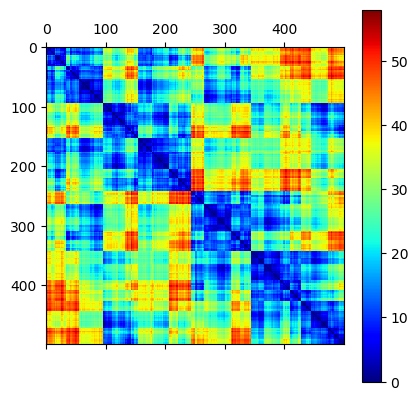}%
\includegraphics[width=2.cm]{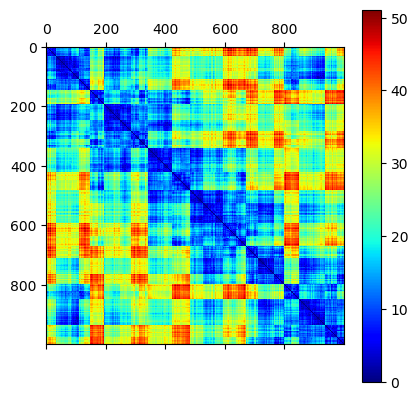} &
\includegraphics[width=2.5cm]{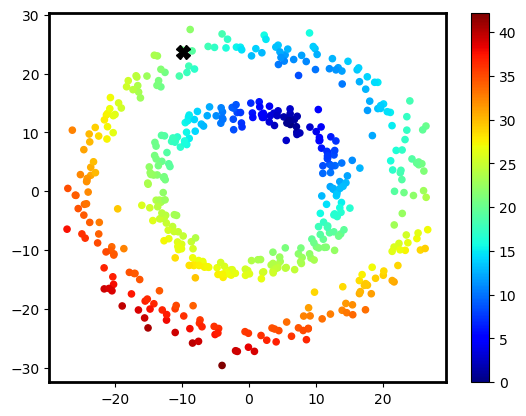}%
\includegraphics[width=2.5cm]{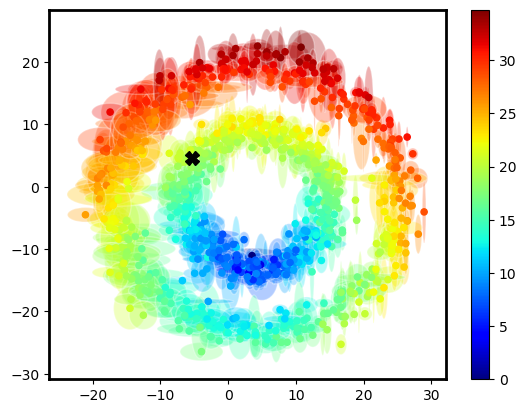}
\end{tabular}
\caption{For each row, from left to right, 1st plot: ground truth points; 2nd plot: Distance matrix for \textsf{STE} embeddings, 3rd plot: Distance matrix for \textsf{ElOE} embeddings, 4th plot: Similarity scores for \textsf{STE} embeddings, Similarity scores for \textsf{ElOE} embeddings.}
\label{fig:toy_dataset_additional}
\end{figure*}

\begin{figure*}[ht!]
\centering
\subfigure[Blobs dataset]{\label{}\includegraphics[scale=0.22]{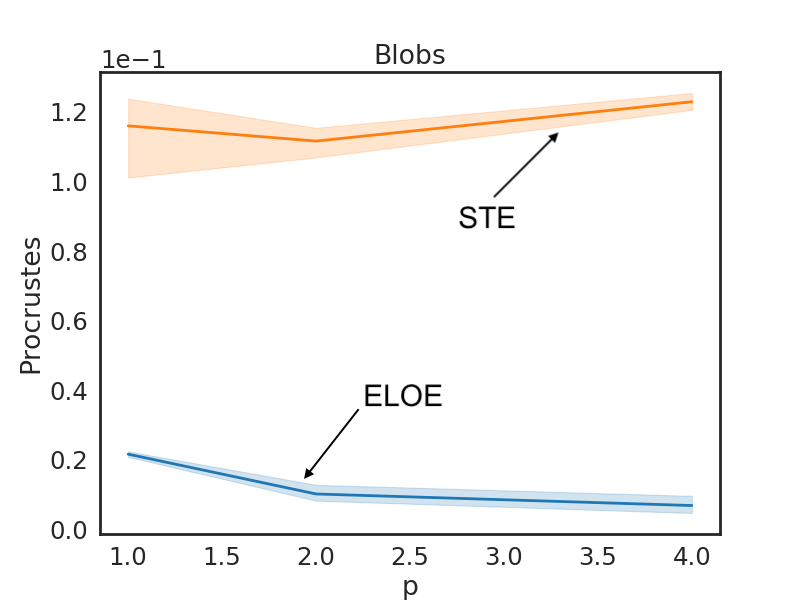}}
\hfill
\subfigure[Moons dataset]{\label{}\includegraphics[scale=0.22]{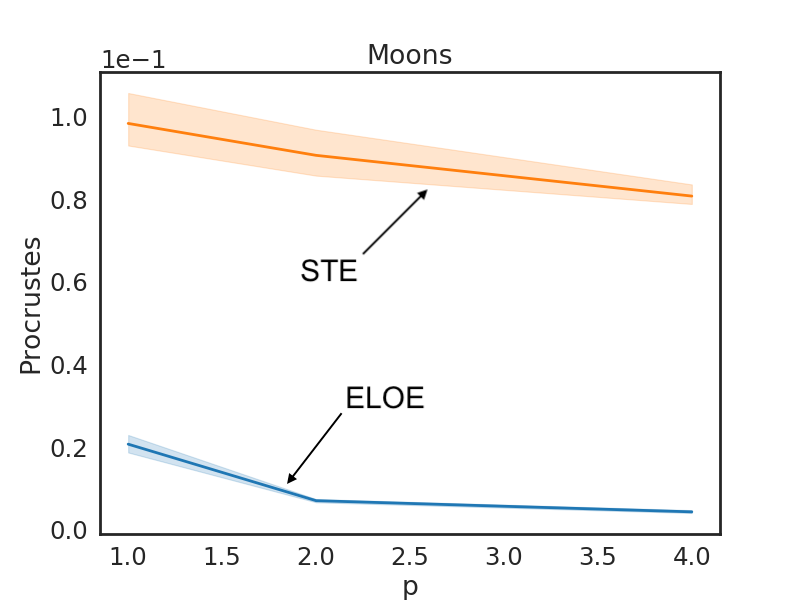}}
\hfill
\subfigure[Circles dataset]{\label{}\includegraphics[scale=0.22]{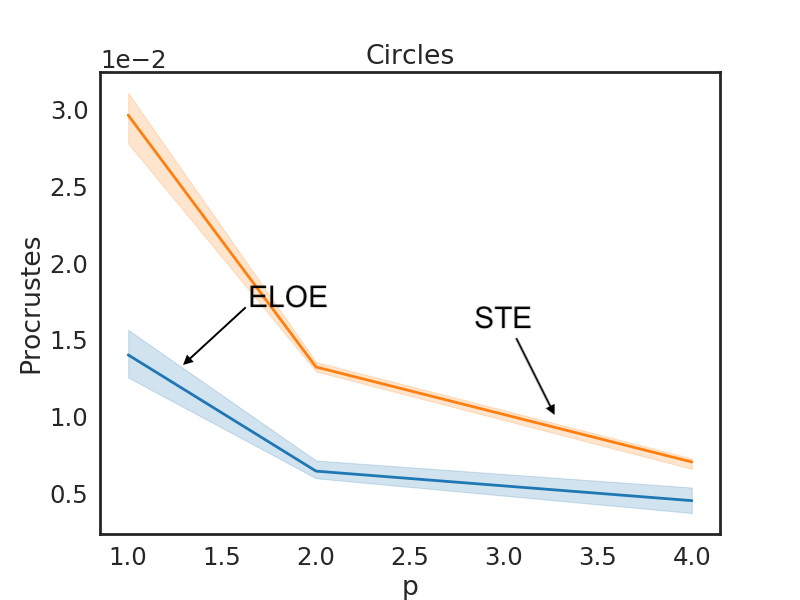}}
\caption{Procrustes distance vs $p$ for synthetic datasets.}
\label{fig:procrustes_vs_p}
\end{figure*}

\begin{figure*}[ht!]
\centering
\subfigure[Blobs dataset]{\label{}\includegraphics[scale=0.22]{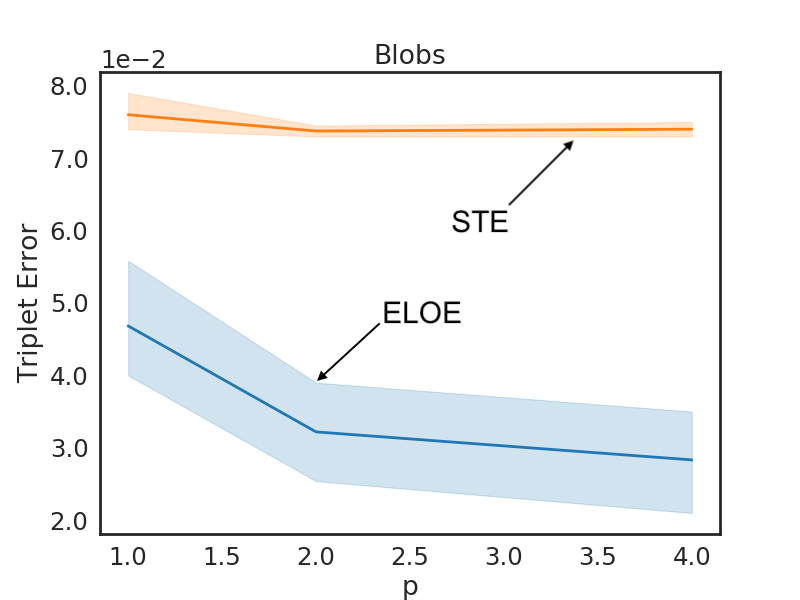}}
\hfill
\subfigure[Moons dataset]{\label{}\includegraphics[scale=0.22]{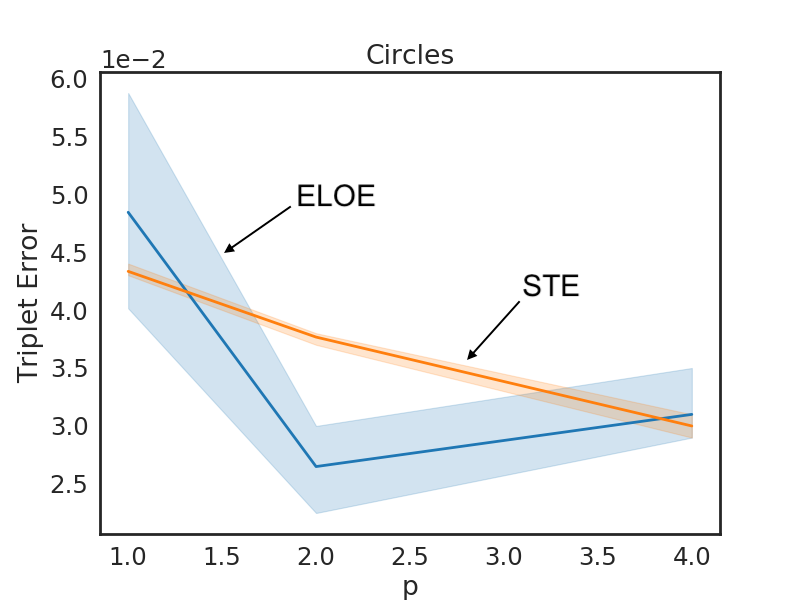}}
\hfill
\subfigure[Circles dataset]{\label{}\includegraphics[scale=0.22]{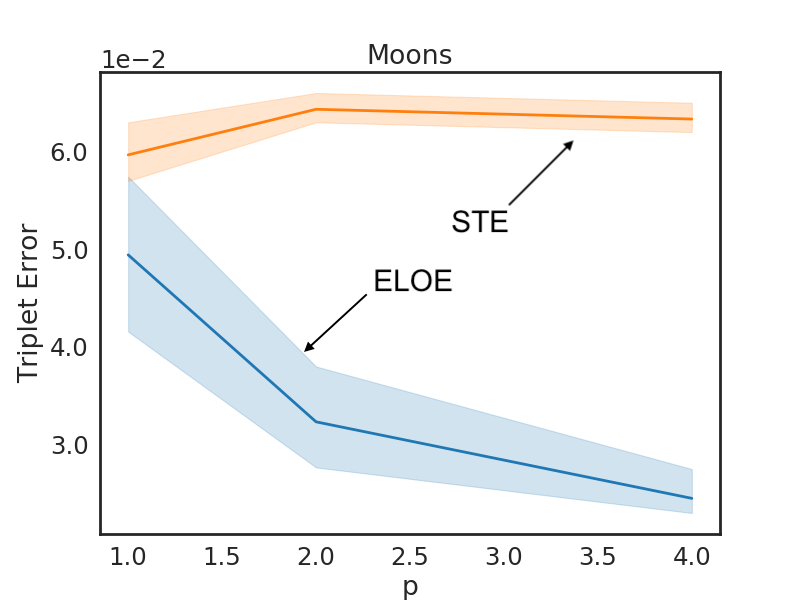}}
\caption{Empirical error vs $p$ for synthetic datasets.}
\label{fig:te_vs_p}
\end{figure*}

In this section, we present additional figures that compare the performance of \textsf{STE} and \textsf{ElOE} embeddings.
Experiments were conducted on three synthetic datasets, each consisting of 1000 points. The results shown in \cref{fig:toy_dataset_additional} are for $p=2$. In particular, the similarity score column underlines the behaviour of the Wasserstein distance. We notice that the similarity is higher between a given point (marked with an "x" in \cref{fig:cifar_graph,fig:voc_graph}) and the others when the distance of their means is lower and the difference of their variance is smaller.
\cref{fig:procrustes_vs_p} and \cref{fig:te_vs_p} illustrates the relation between Procrustes distance and empirical triplet error and the number of triplets, for \textsf{STE} and \textsf{ElOE}. We notice that overall, \textsf{ElOE} performs better than the baseline embeddings. 

\vspace{1cm}

\begin{figure*}[h!]
\centering
\subfigure[]{\label{fig:cifar_big}\includegraphics[scale=0.40]{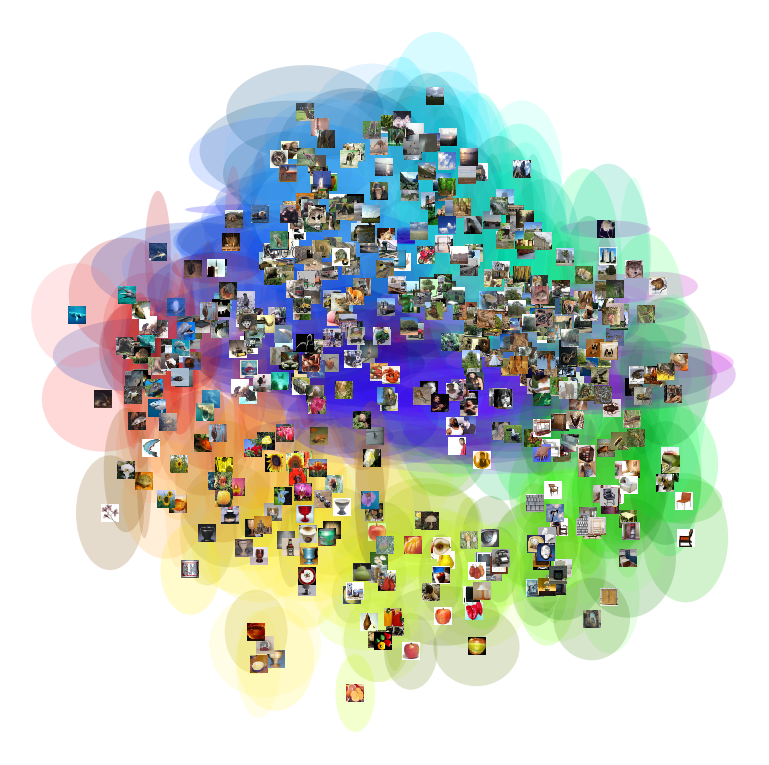}}
\subfigure[]{\label{fig:voc_big}\includegraphics[scale=0.4]{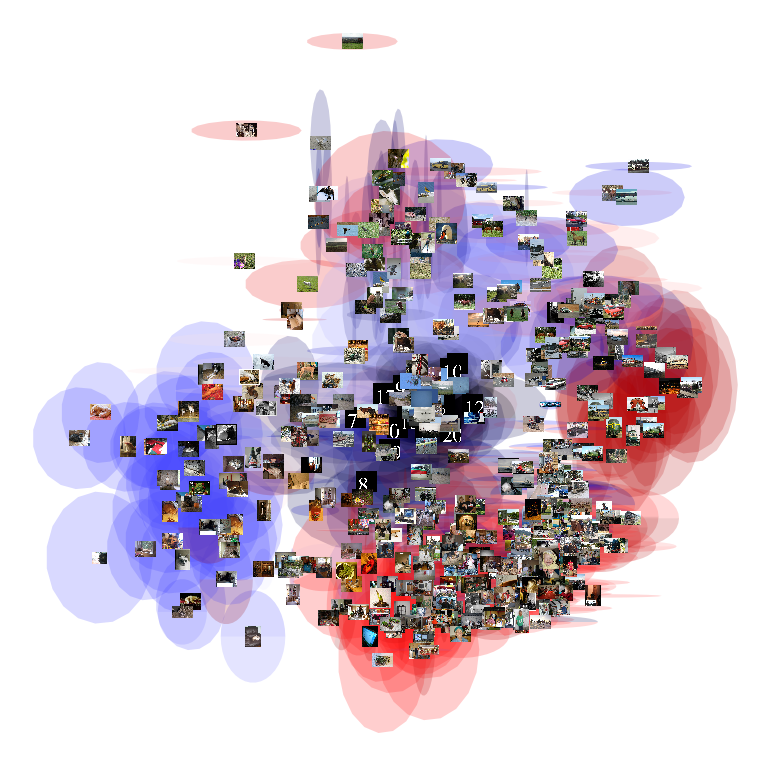}}
\caption{Visualization of the semantic embeddings. (a) CIFAR, (b) VOC. }
\label{fig:graph_big}
\end{figure*}


\end{document}